\documentclass[acmsmall,nonacm]{acmart}

\DeclareUnicodeCharacter{2212}{-}

\usepackage{subcaption}
\usepackage{soul}
\usepackage{amsmath}
\usepackage{multirow}
\usepackage{graphicx}
\usepackage{wrapfig}
\usepackage{bbm}
\AtBeginDocument{%
  \providecommand\BibTeX{{%
    \normalfont B\kern-0.5em{\scshape i\kern-0.25em b}\kern-0.8em\TeX}}}

\setcopyright{acmlicensed}
\copyrightyear{2018}
\acmYear{2018}
\acmDOI{XXXXXXX.XXXXXXX}

\begin{document}

\title[LLM as Virtual Annotators]{Evaluating Large Language Models as Virtual Annotators for Time-series Physical Sensing Data}

\author{Aritra Hota}
\email{aritrahota@kgpian.iitkgp.ac.in }
\affiliation{%
  \institution{Indian Institute of Technology}
  \country{Kharagpur, India}
  \postcode{721302}
}

\author{Soumyajit Chatterjee}
\orcid{0000-0001-5604-2267}
\affiliation{%
  \institution{Nokia Bell Labs}
  \city{Cambridge}
  \country{United Kingdom}
  \postcode{CB3 0FA}
}
\email{soumyajit.chatterjee@nokia-bell-labs.com}

\author{Sandip Chakraborty}
\orcid{0000-0003-3531-968X}
\affiliation{%
  \institution{Indian Institute of Technology}
  \country{Kharagpur, India}
  \postcode{721302}
}
\email{sandipc@cse.iitkgp.ac.in}

\renewcommand{\shortauthors}{Hota et al.}

\begin{abstract}
Traditional human-in-the-loop-based annotation for time-series data like inertial data often requires access to alternate modalities like video or audio from the environment. These alternate sources provide the necessary information to the human annotator, as the raw numeric data is often too obfuscated even for an expert. However, this traditional approach has many concerns surrounding overall cost, efficiency, storage of additional modalities, time, scalability, and privacy. Interestingly, recent large language models (LLMs) are also trained with vast amounts of publicly available alphanumeric data, which allows them to comprehend and perform well on tasks beyond natural language processing. Naturally, this opens up a potential avenue to explore the opportunities in using these LLMs as virtual annotators where the LLMs will be directly provided the raw sensor data for annotation instead of relying on any alternate modality. Naturally, this could mitigate the problems of the traditional human-in-the-loop approach. Motivated by this observation, we perform a detailed study in this paper to assess whether the state-of-the-art (SOTA) LLMs can be used as virtual annotators for labeling time-series physical sensing data. To perform this in a principled manner, we segregate the study into two major phases. In the first phase, we investigate the challenges an LLM like GPT-4 faces in comprehending raw sensor data. Considering the observations from phase 1, in the next phase, we investigate the possibility of encoding the raw sensor data using SOTA SSL approaches and utilizing the projected time-series data to get annotations from the LLM. Detailed evaluation with four benchmark HAR datasets shows that SSL-based encoding and metric-based guidance allow the LLM to make more reasonable decisions and provide accurate annotations without requiring computationally expensive fine-tuning or sophisticated prompt engineering.
\end{abstract}



\received{20 February 2007}
\received[revised]{12 March 2009}
\received[accepted]{5 June 2009}

\maketitle

\section{Introduction}
Human activity recognition (HAR) has been ubiquitous nowadays with myriad applications in domains ranging from personal healthcare monitoring, sports performance analysis, and intelligent living assistance, among others. Traditional activity and context-sensing applications primarily rely on accurate labels to train robust supervised machine learning (ML) models. Conventional methods for obtaining high-quality labels depend on human-in-the-loop-based approaches assisted by sophisticated techniques like active learning~\cite{hossain2019active} and experience sampling~\cite{laput2019sensing}. Notably, one of the primary factors behind involving humans in the labeling process is to create the natural transfer of knowledge from humans to the final ML models. Barring some inherent mislabeling problems, missing labels, and label jitter~\cite{kwon2019handling,zeni2019fixing}, the human-in-the-loop approach often provides accurate labels. This is mainly because the human annotators if chosen correctly, have significant knowledge regarding the activities or the locomotion involved.
\begin{figure}
    \centering
    \includegraphics[width=0.55\columnwidth,keepaspectratio]{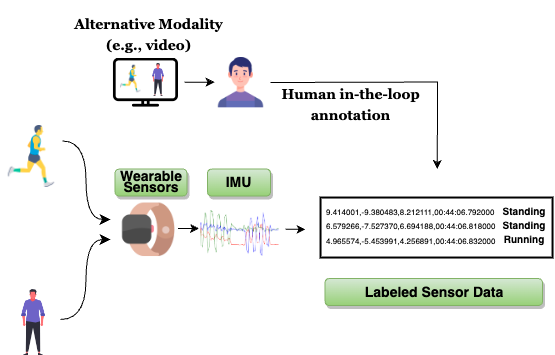}
    \caption{Traditional human-in-the-loop annotation scheme}
    \label{fig:human_in_loop}
\end{figure}

Interestingly, humans cannot perceive the raw physical sensor data, such as the IMU data, as collected from a body-worn device like a smartwatch, a smart glass, or an earable for activity recognition. Therefore, the conventional method is to use one of the alternate modalities from the environment, like video or audio, and use this as an input context to the human annotator for obtaining the labels~\cite{chatterjee2023acconotate} (see~\figurename~\ref{fig:human_in_loop}). Although the existing literature uses human-generated annotations obtained from such alternate modalities as the \textit{gold standard} to train and test models developed over physical sensing (i.e., IMU) data, they often come with several challenges and implementation limitations. (1) The data obtained from the alternate modality (like the video/audio) needs to be perfectly time-synced with the primary modality, i.e., the IMU; otherwise, the annotated labels might not align with the corresponding activities. Although existing approaches try to handle such synchronization issues~\cite{zhang2020syncwise}, perfectly aligning the activity boundaries in the continuous activity space is severely challenging. (2) There might be perceptional bias from the annotators when the visible activity signatures have only some minute differences in the auxiliary modality (say, \textit{running} vs. \textit{jogging}), whereas the physical sensing data have significant differences (i.e., \textit{running} introduces higher intensity than \textit{jogging} in the acceleration signatures). Considering these challenges, we ask the following question in this paper: \textit{Rather than depending solely on the auxiliary modality for ground truth annotation, can we augment the annotation process with information extracted from the physical sensing data to avoid human-induced bias, as discussed above?}


Interestingly, most recent large language models (LLMs) have also been trained with vast knowledge from publicly available data, potentially containing information from numeric data beyond the usual text data~\cite{liu2023large,yao2022react}. This, in turn, opens up an avenue to investigate whether these LLMs can directly annotate the raw sensor data without using alternate modalities. To assess this, we first set up a detailed pilot study (Section~\ref{motivation}) whereby we analyze whether the state-of-the-art (SOTA) LLMs like GPT-4 can quickly identify the labels (the activity classes) in a multi-class classification setup. This study shows that although GPT-4 can understand the accelerometer signatures, it fails to separate them across different classes even with a fixed number of bootstrap samples as example observations. Notably, some recent studies like~\cite{liu2023large} have already attempted to investigate how LLMs can be used with sensor data streams and reached similar conclusions.

One important alternative to resolving this concern can be efficient fine-tuning of the LLMs to achieve higher accuracy in the downstream task of recognizing the activity classes~\cite{zhang2023balancing}. However, fine-tuning these large models with a billion parameters can be challenging given their compute and data requirements~\cite{gao2024llm}. To avoid this, a few works like~\cite{kim2024health} have also looked into prompt-tuning as a viable alternative where the main objective is to include more context information as a part of the prompt, which in turn could provide the LLM with some additional information regarding the data. Although this prompt tuning approach has been seen to work well with specialized medical sensors~\cite{kim2024health}, images~\cite{qu2023layoutllm}, and videos~\cite{zhang2023video}, they have often failed to achieve the desired accuracy for locomotive sensor data~\cite{jin2024position} due to its inherent lack of enough context information. 
\begin{figure}
    \centering
    \includegraphics[width=0.90\columnwidth,keepaspectratio]{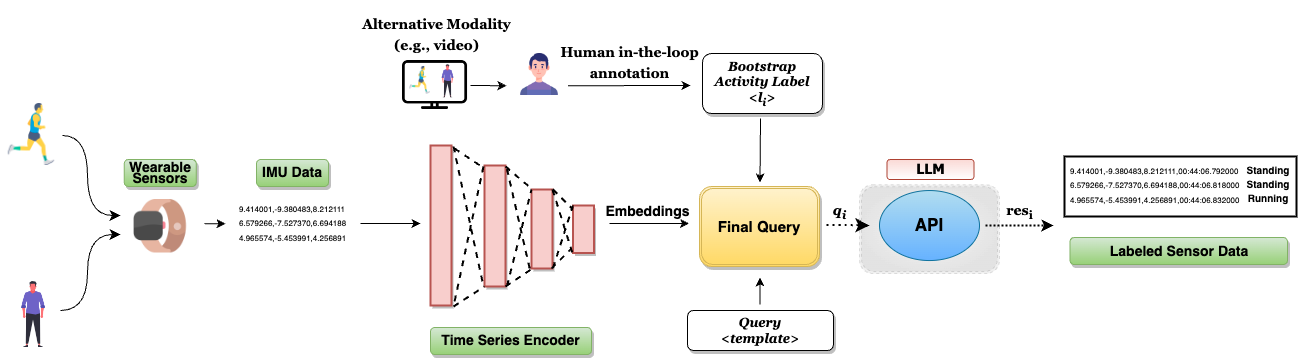}
    \caption{Proposed setup for evaluating LLMs with encodings obtained from an encoder pre-trained self-supervised manner without access to labeled data (Phase 2).}
    \label{fig:proposed_setup}
\end{figure}

Interestingly, works like~\cite{liang2022holistic,belyaeva2023multimodal} have investigated a different approach to designing specialized encoders that can provide meaningful representations to the LLMs, which then can be interpreted by the LLM to offer accurate responses to the downstream task. However, unlike for modalities like images, IMU data is often marred with noises and body-position-based domain heterogeneity~\cite{jain2022collossl}. This, in turn, has led to the unavailability of robust pre-trained encoders for locomotive data, which, in particular, is more challenging when no labels are available and we rely solely on the LLM for annotating the raw data.

Notably, papers like~\cite{tang2020exploring,ramamurthy2022cogax, jain2022collossl} have shown that self-supervised pre-training can create robust encoders which, when fine-tuned with a small amount of bootstrap data, can achieve significant accuracy. More specifically, this pre-training step often exploits the idea of contrastive learning~\cite{tang2020exploring, jain2022collossl}, which allows the encoder to learn robust representations without requiring the labeled data. Motivated by this idea, we, in this paper, design a study to evaluate LLMs with inertial data encoded using a self-supervised pre-trained encoder with a supporting prompt that includes a few bootstrap examples (see \figurename~\ref{fig:proposed_setup}). The principled evaluation of four benchmark human activity recognition datasets highlights the overall consistency and boost in accuracy of the annotations provided by GPT-4 without any need for fine-tuning or labeling data. We also highlight the cost and time required for using LLMs as virtual annotators. This shows that as virtual annotators, LLMs can help us reduce the overall cost and time for human annotation and eliminate the cost associated with recording and storing data from other alternate modalities while reducing the risk of privacy leaks. Finally, we discuss some of the critical future directions that can be explored given the future use of LLMs as virtual annotators and how such systems can be standardized in the general pipeline of developing sophisticated ML applications and services. In summary, the significant contributions of this paper are as follows. 
\begin{enumerate}
    \item \textbf{Exploring LLM's Capability as Virtual Annotators of the Physical Sensing Data:} We perform a thorough analysis to explore whether SOTA LLMs, like GPT-4, can work as virtual annotators to label physical sensing data, with locomotive data as a use-case for the downstream task of human mobility annotation. We show that although LLMs can understand locomotive data like a triaxial accelerometer signature, they fail to properly annotate the data with the corresponding mobility classes, even when sample bootstrap examples are provided along with the contextual information in the prompt. 
    \item \textbf{Extending LLM's Capability as Virtual Annotators of the Physical Sensing Data:} Based on the findings from the pilot study, we develop an approach to encode the locomotive data in an embedding space that inherently learns the signatures of different classes and thus provides rich contextual information to the LLMs for annotating the accelerometer signatures with the latent embedding of the raw locomotive data. We explore self-supervised contrastive learning with two different setups for this purpose -- (a) with time-domain signatures only and (b) with both time and frequency-domain signatures. To the best of our knowledge, this is the first work that helps LLMs enhance their capability for annotating locomotive data from the raw accelerometer with the corresponding human mobility classes without requiring any retraining.
    \item \textbf{Thorough Evaluation, Analysis, and Discussion:} We perform a detailed investigation using four benchmark HAR datasets on how the encoded accelerometer data offers better reasoning for the LLM to provide accurate annotations with the increasing number of examples. Notably, with paid LLMs like ChatGPT-4, a mere performance assessment with accuracy as the key highlight is not sufficient as there are factors like the number of tokens and rate limits are restrictions that impact the final usage of such a service for annotating a dataset. Understanding this, we in this paper, also assess the time and cost of analysis, summarise the trade-offs, and discuss the advantages of using LLMs as a virtual annotator for time-series physical sensing data. Additionally, we also investigate the inherent limitations of using an unimodal sensor data from a triaxial accelerometer with a comprehensive analysis considering more than two classes.
\end{enumerate}

The rest of the paper is organized as follows. Section~\ref{rel_work} provides a detailed study of the related literature in exploring LLMs for different practical use cases of time-series data and using self-supervised models for time-series data analysis. Section~\ref{datasets} briefly discusses the datasets used for this study. We next perform a thorough pilot study to analyze LLM's capability in annotating time-series locomotive data, as mentioned in Section~\ref{motivation}. Based on the findings, we next explore in Section~\ref{system} the use of self-supervised contrastive learning to encode the locomotive data in an embedding space, which can help LLMs annotate the data correctly. Section~\ref{eval} provides the implementation details and analyzes the empirical results to highlight the significant observations. We discuss our broad learning and the critical insights from this study in Section~\ref{discuss}. Finally, Section~\ref{conclusion} concludes the paper.

\section{Related Work}
\label{rel_work}
Before diving deeper into the study's design, we first perform a detailed survey on the recent works investigating the applications surrounding LLMs considering time-series data.
\subsection{Using LLMs for Understanding Time-Series Data}
One of the most critical evaluations that SOTA LLMs have faced is towards understanding of data modalities other than the natural language. Works like ELIXR~\cite{xu2023elixr}, multimodal approaches like Med-PaLM~\cite{tu2024towards}, and HeLM~\cite{belyaeva2023multimodal} have explored using LLMs with different other data modalities like images and tabular data. Interestingly, unlike image or text data, time-series data is often much more challenging to interpret~\cite{jin2024position}. Naturally, it is crucial to investigate whether SOTA LLMs, trained on a massive volume of alphanumeric data, can comprehend time-series data. Subsequently, works like~\cite{liu2023large,xue2023promptcast,kim2024health,zhang2024large,xu2023penetrative,jin2023time} have started looking into how LLMs can assist in solving tasks that involve time-series data. Notably, many of these works used examples to provide more context to the LLM to gain assistance in predicting the final output.

Interestingly, works like~\cite{liu2023large} used the raw sensor data as an example; however, the idea did not show consistent performance with increasing classes. More recent works, like~\cite{brown2020language,kim2024health,xue2022leveraging,xu2023penetrative} have attempted to provide more \textit{guidance and examples} through \textit{in-context learning} which allows the inclusion of more information within the prompt and does not necessarily need any fine-tuning of the LLM parameters. This approach also has the immense benefit of allowing the transfer of human expertise to the LLM~\cite{xu2023penetrative,stanford_incontext}. Furthermore, works like~\cite{kim2024health,xu2023penetrative} also show the importance of providing signal-processing~\cite{xu2023penetrative} or interpretation-based guidance~\cite{kim2024health} which allows the model to make predictions regarding the query sample. However, one key aspect that we observe in most of these aforementioned works is that in most cases, the input time-series is from specialized sensors like medical~\cite{kim2024health} or temperature sensors~\cite{xu2023penetrative} for which expert guidance is possible. For example, the prompt can provide what a typical high heart rate means and how it can be observed in the ECG signal. However, with more generalized sensors like accelerometers, it is hardly possible to provide some expert guidance directly, given the inherent nature of the sensor data itself. To mitigate this, works like~\cite{jin2024position} have discussed the idea of using time-series encoders, albeit with a proposal to align the time-series features with language-specific representations, which in turn would need further fine-tuning of the LLM. In this study, however, we explore using a time-series encoder to provide more context to the LLM without fine-tuning. Based on this objective, we assess whether an LLM like GPT-4 can be used as a service to obtain annotations for the unlabeled accelerometer data for HAR.  
\subsection{Self-supervised Contrastive Learning with Time-series Data}
Human-in-the-loop annotation scheme is expensive, inefficient~\cite{wu2022survey}, and also known to be noisy~\cite{zeni2019fixing,kwon2019handling}. Owing to these concerns in the recent past, there have been a plethora of works that have looked into self-supervised learning (SSL), which proposes the development of label-efficient models that explore the patterns present inside the data to generate robust representations~\cite{chen2020simple}. These representations can eventually be used to fine-tune a classifier with a small amount of labeled data~\cite{chen2020simple}. More specifically, works like SimCLR~\cite{chen2020simple} and SwAV~\cite{caron2020unsupervised} explored the idea of contrastive learning where the objective is to train an encoder that can cluster similar samples whereas as separate diverse samples in the embedding space without the need to access labeled data. To achieve this with image data, these works often relied on different augmentation approaches like cropping, rotations, adding noise, scaling, etc., to generate augmented versions of the same data and then use them to train the encoder in a self-supervised manner without labels~\cite{jaiswal2020survey}.

Although originally explored for images, recent works like~\cite{ramamurthy2022cogax,tang2020exploring,jain2022collossl,deldari2021time,deldari2022cocoa,zhang2022self,xu2023retrieval} have extended this idea to sensor time-series data. Most of these works explored different ideas for performing contrastive learning by doing time-series specific augmentations like jitter, noise addition, etc.~\cite{tang2020exploring}, whereas works like~\cite{jain2022collossl} exploited the multi-device setting to find similar and dissimilar samples. Interestingly, works like~\cite{zhang2022self} and~\cite{xu2023retrieval} have revisited the approach by looking into frequency and subsequence (or motifs) based information as well. Notably, the overall approach of using an encoder pre-trained using the self-supervised contrastive learning-based approach is extremely efficient and robust, especially for wearable sensing-related tasks~\cite{haresamudram2022assessing,qian2022makes,spathis2022breaking,dhekane2023much,yuan2022self}. Motivated by these observations and the potential of SSL in a contrastive framework, we in this paper analyze the possibility of using encoded accelerometer data for obtaining high-quality annotations from SOTA LLMs without the need for computationally expensive fine-tuning of the LLM or sophisticated prompt engineering.
\section{Datasets Used for the Analysis}
\label{datasets}
From the detailed related work on analyzing time-series data using LLMs and a background study on SSL, mainly focusing on HAR, we chose four benchmark datasets for the studies conducted in the remainder of this paper. Before we move into the details of the studies, in this section, we summarize the dataset details and how we use them in this study. A brief description of the datasets can be seen in \tablename~\ref{tbl:datasets}, and the details follow.
\begin{table}[]
\centering
\scriptsize
\caption{Details of the dataset used in this analysis. In this setup, we hold out a set of participants for generating the test queries for evaluating the LLMs performance as a virtual annotator.}
\label{tbl:datasets}
\begin{tabular}{|l|l|l|l|l|l|}
\hline
\textbf{Dataset} & \textbf{\begin{tabular}[c]{@{}l@{}}\#Training\\ Samples\end{tabular}} & \textbf{\begin{tabular}[c]{@{}l@{}}\#Testing\\ Samples\end{tabular}} & \textbf{Selected Device Position} & \textbf{Chosen Classes}   & \textbf{\begin{tabular}[c]{@{}l@{}}Sampling\\ Frequency\end{tabular}} \\ \hline
MotionSense~\cite{malekzadeh2019mobile}      & 337171                                                                & 72728                                                                & Trousers' front pocket   & Jogging \& Upstairs       & 50 Hz                                                                 \\ \hline
PAMAP2~\cite{misc_pamap2_physical_activity_monitoring_231}           & 1931748                                                               & 509695                                                               & Hand                     & Running \& Walking        & 100 Hz                                                                \\ \hline
UCI HAR~\cite{misc_human_activity_recognition_using_smartphones_240}          & 164718                                                                & 47033                                                                & Waist                    & Standing \& Walk upstairs & 50 Hz                                                                 \\ \hline
HHAR~\cite{misc_heterogeneity_activity_recognition_344}            & 3020605                                                               & 521516                                                               & Arm                      & Standing \& Stairs up     & 50$\sim$200 Hz                                                        \\ \hline
\end{tabular}
\end{table}
\begin{enumerate}
\item{\textbf{MotionSense --}} The~\textit{MotionSense dataset} \cite{malekzadeh2019mobile} consists data from 24 participants performing 6 different activities (Jogging, Sitting, Standing, Walk upstairs, Walk downstairs and Walking). The data was captured from the embedded IMU sensors (only the reading from triaxial accelerometer) of a smartphone which was kept in the trousers' front pocket of the participants. The accelerometer readings were sampled at a frequency of 50Hz. Notably, the dataset is divided into 16 different trails and we considered the last trial for each activity for the analysis. We have considered ``Jogging'' and ``Upstairs'' as the activity classes and subjects 1, 6, 14, 19 and 23 as our test subjects. The rest of the subjects across all the activity classes constitute the training set.

\item{\textbf{PAMAP2 -- }}
The~\textit{PAMAP2 dataset} \cite{misc_pamap2_physical_activity_monitoring_231} comprises a total of 9 participants performing 18 different physical activities with 6 optional activities. Out of these $24$ activities we have considered the 12 primary activities (Ascending stairs, Cycling, Descending stairs, Ironing, Lying, Nordic walking, Rope jumping, Running, Sitting, Standing, Vacuum cleaning and Walking). While collecting this dataset, the participants were equipped with 3 IMU's each having a sampling rate of 100Hz wirelessly attached to 3 body positions: head, ankle and chest. However, in our evaluation, we have considered the data captured using the IMU sensor worn on the hands only. Additionally, we have considered ``Running'' and ``Walking'' as the activity classes and subjects 101 and 108 as the test subjects. The rest of the subjects belonging to all the 12 activity classes comprise the training set.

\item{\textbf{UCI HAR --}}
The \textit{UCI HAR dataset}~\cite{misc_human_activity_recognition_using_smartphones_240} contains inertial data for 6 different activities (Laying, Sitting, Standing, Walking, Walking downstairs and Walking upstairs). These activities were performed by wearing a smartphone around their waist. The IMUs have a sampling rate of 50Hz. For our setup, we have chosen ``Standing'' and ``Walking upstairs'' as the activity classes and subjects 1, 3, 5, 7, 9, 11, 13, 15 and 17 as the test subjects whereas, all other remaining subjects across all activity classes are part of the training set.

\item{\textbf{HHAR --}}
The \textit{HHAR dataset}~\cite{misc_heterogeneity_activity_recognition_344} contains data recorded from 9 participants performing 6 activities (Biking, Sitting, Standing, Stair down, Stair up and Walking). The participants were given 8 smartphones and 4 smartwatches. For this study, we have used the accelerometer readings from the smartwatch worn on one hand of the user. The accelerometer readings were sampled at a frequency of 50-200 Hz depending on the device used. We have considered ``Standing'' and ``Stair up'' as our activity classes and subjects 3 and 9 as our test subjects while considering all other remaining participants across all activities for the training set.
\end{enumerate}
\section{Phase 1: Raw Sensor Data for Virtual Annotation}
\label{motivation}
\begin{figure}
    \centering
    \includegraphics[width=0.9\columnwidth,keepaspectratio]{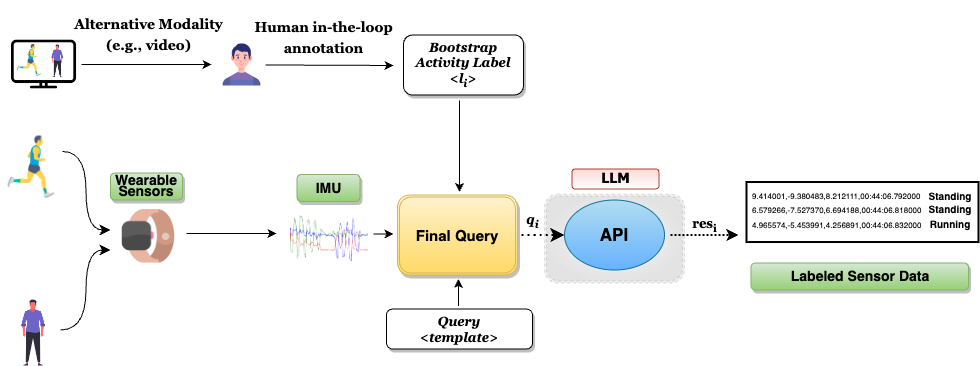}
    \caption{Proposed setup for the motivational experiments (Phase 1)}
    \label{fig:motivation_setup}
\end{figure}
Before we proceed in analysing how different encoding schemes can impact the understanding of sensor data by the SOTA LLMs, we first investigate what are the key scenarios where the SOTA LLMs fail or succeed in providing quality labels when the given input is the raw inertial data. As shown in~\figurename~\ref{fig:motivation_setup}, this is significantly different from the traditional human-in-the-loop annotation scheme as the LLM is directly provided with the raw sensor data samples instead of any alternative modality like video or audio from the environment. The details of the pilot experiments conducted in this phase, the basic setup and the key takeaways are summarized as follows.
\subsection{Design of the Pilot Study}
To assess the potential of LLMs as virtual annotators, we start by designing a set of experiments. We choose the PAMAP2~\cite{misc_pamap2_physical_activity_monitoring_231} as described in Section~\ref{datasets}. We have used ChatGPT-4 (or GPT-4) as the representative LLM in the remainder of the paper to highlight the idea of using a paid black-box LLM for getting annotations. The details of the experimental setup follow.



\subsubsection{Setup} For the analysis, (a) we first choose the simplest \textit{binary class annotation task} with data selected for the ground-truth activity classes of \textit{running} and \textit{walking}. (b) Next, we assess the quality of labels generated by the LLM by providing some \textit{bootstrap examples} containing \textit{sensor data with ground-truth labels} sampled \textit{across all subjects}. A visual representation of the setup has been shown in \figurename~\ref{fig:motivation_setup}. 

\subsubsection{Queries}
\label{query_motiv}
For obtaining the labels as a response from the LLM, we query it using a set of designed queries, including the raw sensor data and the body position-based information. The template for the queries we use is as follows.\\

\noindent
\textit{``Classify the following triaxial accelerometer data in meters per second square as either walking or running provided that this data is coming from the wearable worn by the user on their dominant hand: [$q_x$ $q_y$ $q_z$]. Answer in one word, either walking or running.''}\\

\noindent
Here, [$q_x$ $q_y$ $q_z$] is the query sensor sample from the triaxial accelerometer. Notably, many recent works like~\cite{brown2020language,liu2023large} often provide examples (few-shots) within the prompt to allow the LLM to gain additional context regarding the data. In this study, we first provide a small amount of bootstrap raw accelerometer data with class labels as examples. To include these examples, we modify the query template by adding the following at the beginning.\\

\noindent
\textit{``Given the following triaxial accelerometer data in meter per second square coming from the wearable worn by the user on their dominant hand corresponds to running : 
[$s_{x_1}$  $s_{y_1}$  $s_{z_1}$]
Given the following triaxial accelerometer data in meter per second square coming from the wearable worn by the user on their dominant hand corresponds to walking : 
[$p_{x_1}$  $p_{y_1}$  $p_{z_1}$]''}\\

\noindent
Here, [$s_{x_1}$  $s_{y_1}$  $s_{z_1}$] and [$p_{x_1}$  $p_{y_1}$  $p_{z_1}$] are the example sensor samples chosen across different subjects for the two different classes. We vary the number of examples to observe the change in the quality of annotations.
\subsection{Key Observations}
\begin{figure*}[]
    \centering
    \begin{minipage}{0.33\textwidth}
        \centering
        \includegraphics[width=\columnwidth,keepaspectratio]{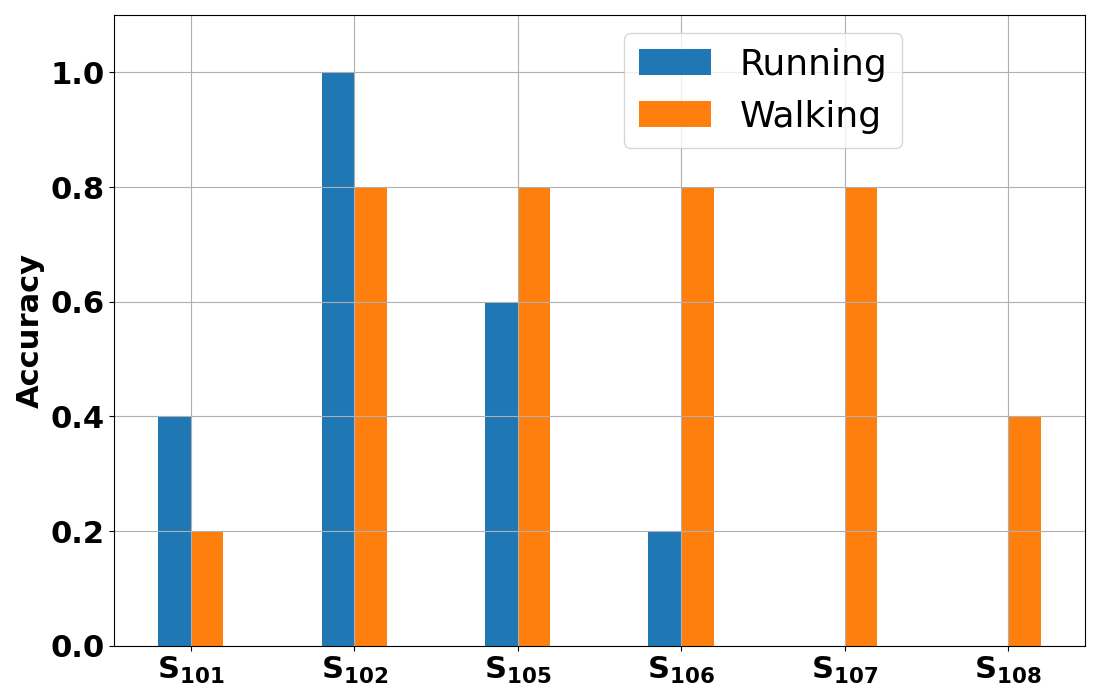}
        \subcaption{}
        \label{fig:cross_user_three_shot}
    \end{minipage}%
    \begin{minipage}{0.33\textwidth}
        \centering
        \includegraphics[width=\columnwidth,keepaspectratio]{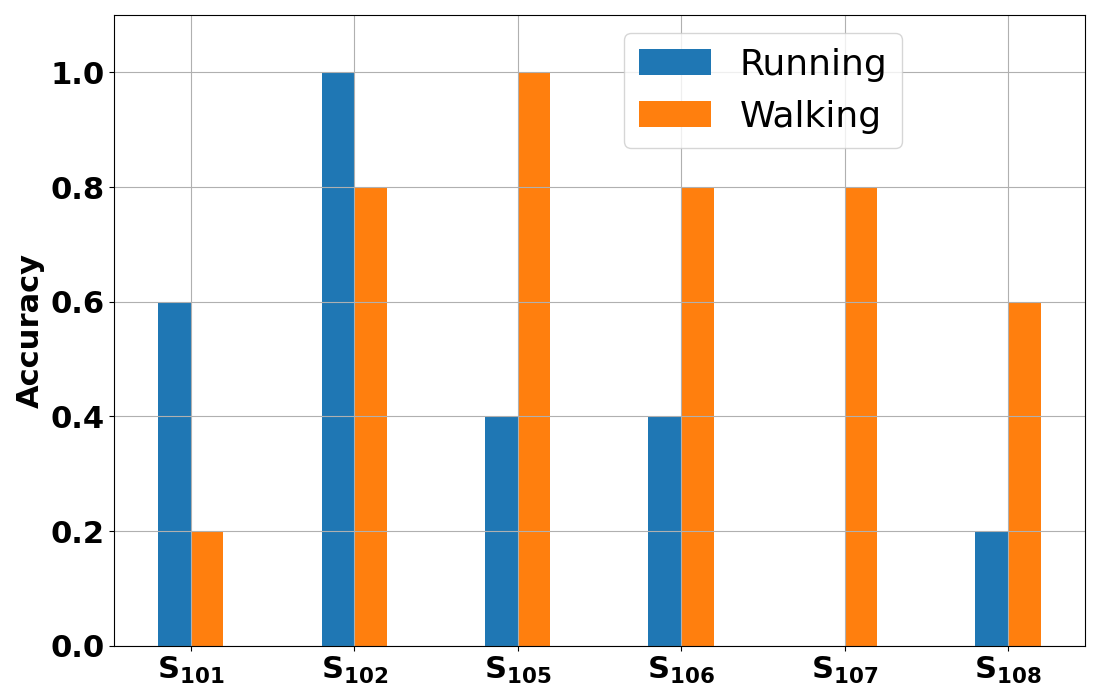}
        \subcaption{}
        \label{fig:cross_user_ten_shot}
    \end{minipage}%
    \begin{minipage}{0.33\textwidth}
        \centering
        \includegraphics[width=\columnwidth,keepaspectratio]{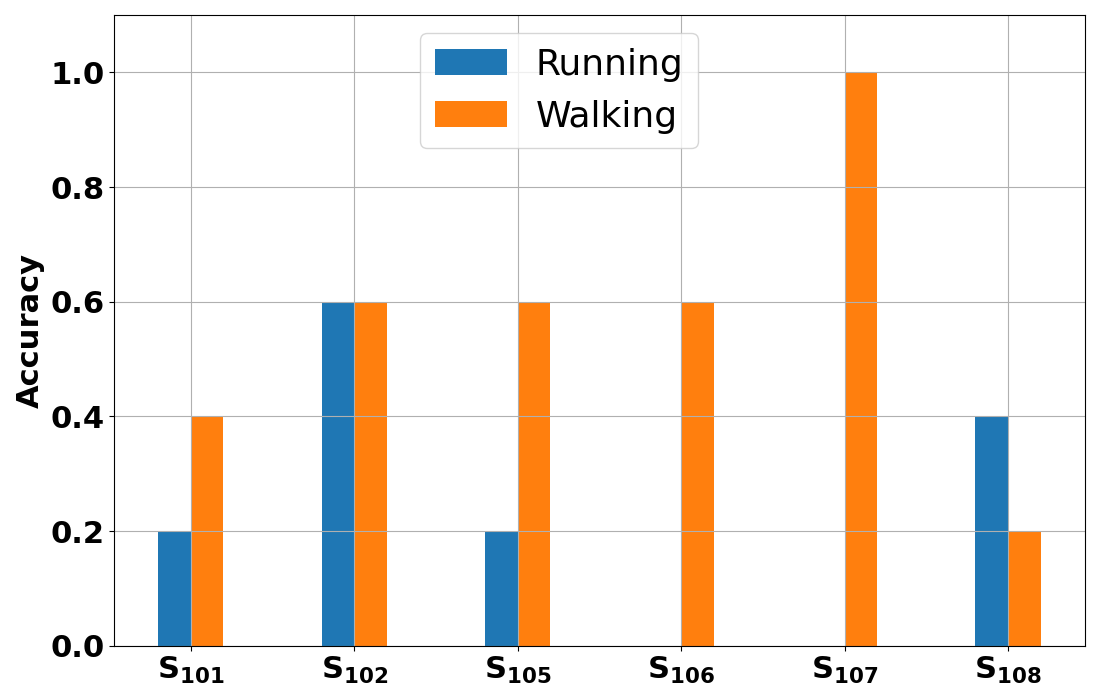}
        \subcaption{}
        \label{fig:cross_user_twentyfive_shot}
    \end{minipage}
    \caption{Subject-wise mapping accuracy in PAMAP2 obtained using GPT-4 with different numbers of examples drawn across different subjects (a) 3 examples, (b) 10 examples, and (c) 25 examples. Only those subjects who have both the ground-truth activity classes present in the dataset have been included in this analysis.}
    \label{fig:consistency_gpt4}
\end{figure*}

\begin{figure*}[]
    \centering
    \begin{minipage}{0.33\textwidth}
        \centering
        \includegraphics[width=\columnwidth,keepaspectratio]{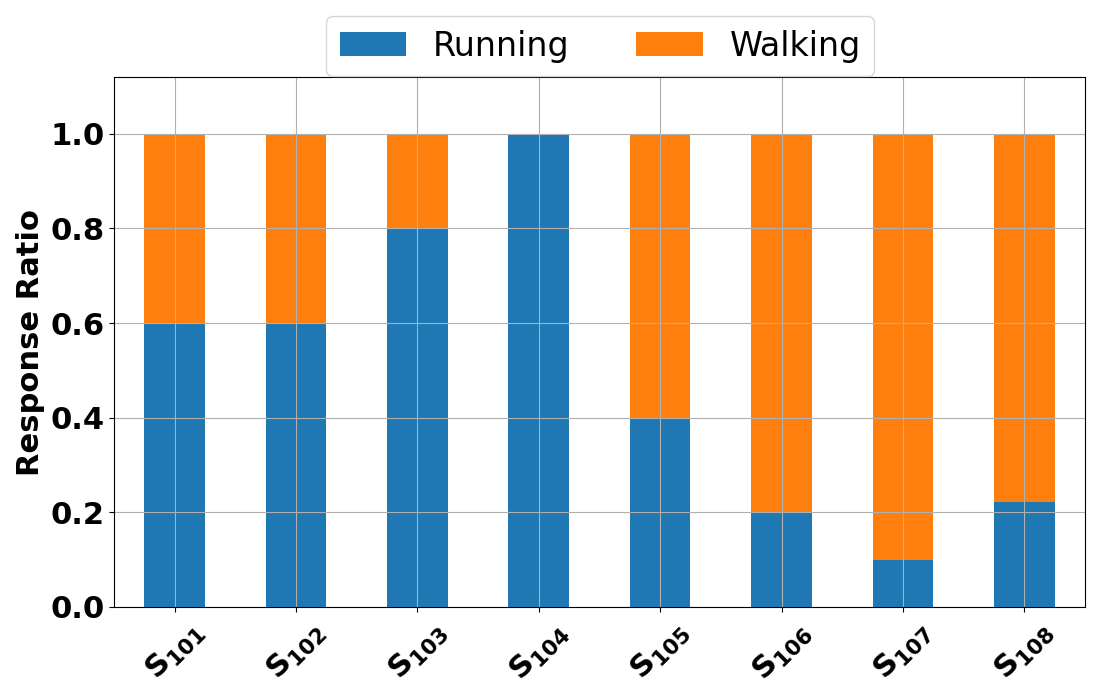}
        \subcaption{}
        \label{fig:cross_user_three_shot_rb}
    \end{minipage}%
    \begin{minipage}{0.33\textwidth}
        \centering
        \includegraphics[width=\columnwidth,keepaspectratio]{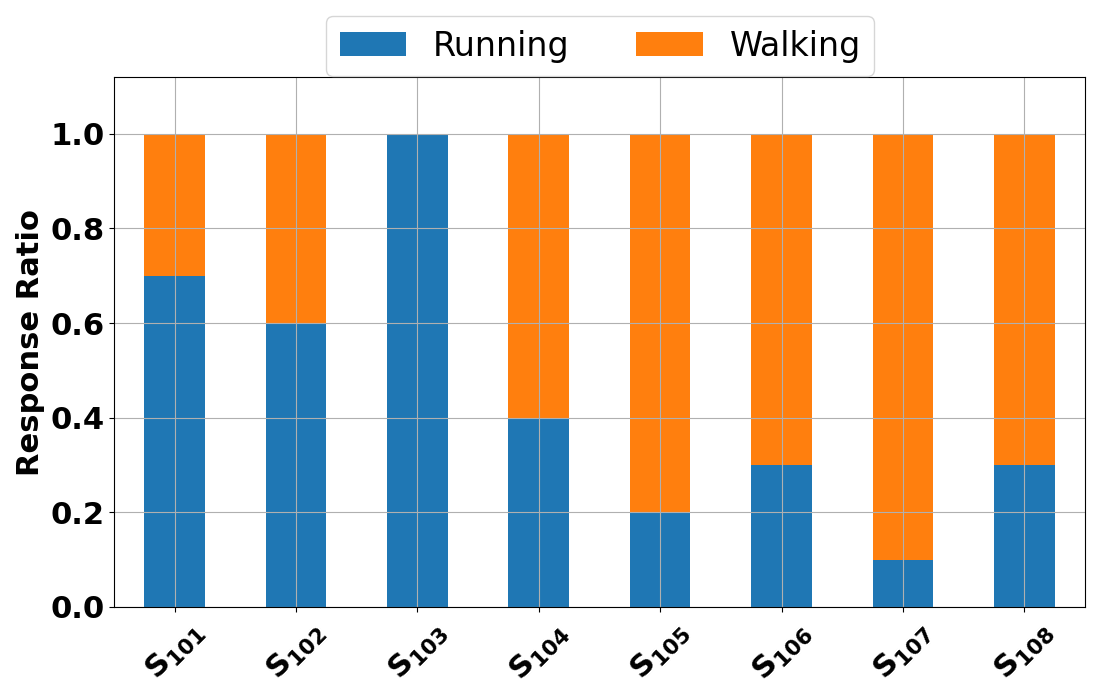}
        \subcaption{}
        \label{fig:cross_user_ten_shot_rb}
    \end{minipage}%
    \begin{minipage}{0.33\textwidth}
        \centering
        \includegraphics[width=\columnwidth,keepaspectratio]{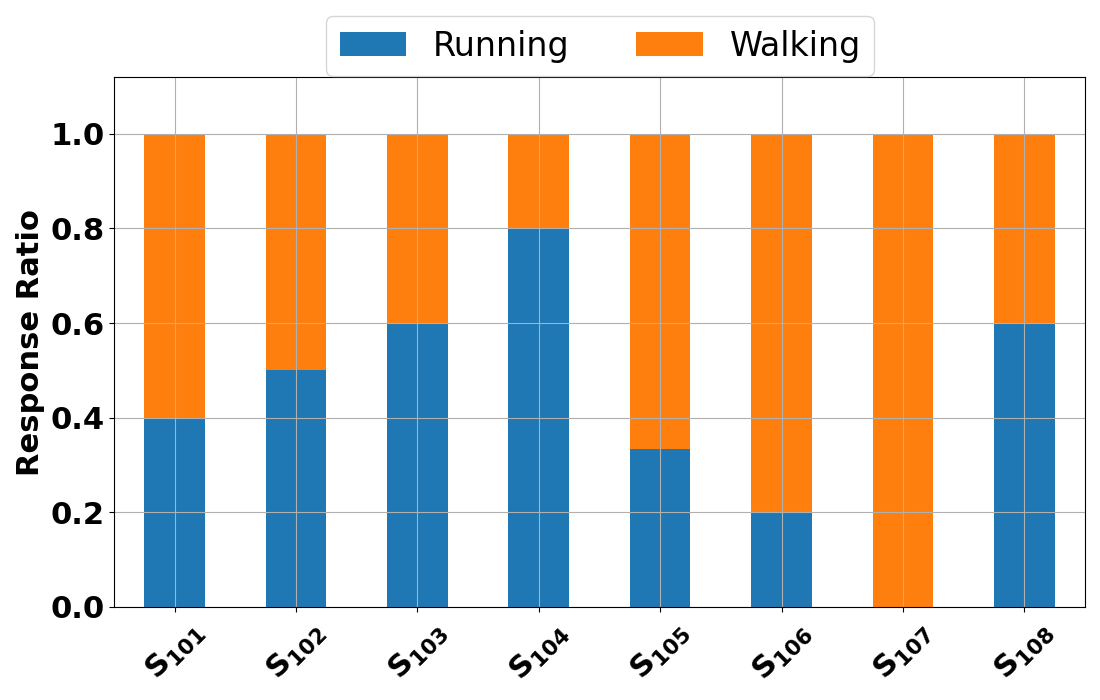}
        \subcaption{}
        \label{fig:cross_user_twentyfive_shot_rb}
    \end{minipage}
    \caption{Response bias exhibited by GPT-4 for the data from PAMAP2 for binary classes with different number of examples drawn across different subjects (a) 3 examples, (g) 10 examples, and (h) 25 examples.}
    \label{fig:bias_gpt4}
\end{figure*}
We next summarize the key observations in terms of accuracy of the responses and also look into whether GPT-4 specifically predicts one class more than the other in a biased manner (a.k.a response bias).

\subsubsection{Mapping Sensor Data to Labels}
We first start by looking at the accuracy of how well the LLM (here GPT-4) can map the raw sensor data to the associated physical activity labels. Interestingly, we see that for \textbf{certain subjects with specific classes, the mapping accuracy is quite appreciable} (see \figurename~\ref{fig:consistency_gpt4}), especially if the task is understanding binary classes only. This further motivated us to see whether this mapping accuracy can be improved by giving the LLMs more examples from the different classes. However, we observed that \textbf{with examples, this mapping accuracy did not change significantly}. In this paper, we investigate this further by -- (a) looking into the consistency of the generated labels with an increasing number of examples and (b) whether the responses themselves are biased based on the numerical values that the LLM is receiving in the query and the examples. The summary of these observations is provided as follows.
\subsubsection{Consistency with Increasing Examples}
One of the primary factors behind relying on human annotators has been consistency. This usually relates to the overall consistent improvement in the quality of the annotations, with an increasing number of bootstrap samples provided as examples during the labeling process. However, in this case, we observe that GPT-4 lacks in terms of consistency. More specifically, we \textbf{do not see any significant correlation} between the \textbf{mapping accuracy and the number of examples} (see \figurename~\ref{fig:consistency_gpt4}).
\subsubsection{Response Bias}
Finally, we also analyzed the responses obtained for the LLM for bias. Notably, we see (from \figurename~\ref{fig:bias_gpt4}) that irrespective of the number of classes or examples provided, the LLM is mostly biased towards labeling majority of the sensor data as `walking'. This actually tells us that without any specialized encoding of the raw sensor data the chances that \textbf{LLMs will map the data to wrong labels} is appreciably high. Additionally, this also highlights that the \textit{significant mapping accuracy} for certain classes might be completely because of this \textit{biased mapping} that the LLM has generated.
\subsubsection{Reasoning}
\label{motiv_reason}
With GPT-4, for majority of the cases, we obtained the following response. For example, for subject 101 with ground-truth `running', one of the responses was as follows.\\

\noindent
\textit{``Sorry, as an AI model, I'm not able to classify real-time activities based on raw accelerometer data instantly. This type of classification typically involves training a machine learning model on large datasets to recognize the patterns associated with different activities. If you've trained such a model, you should input this data there.''}\\

\noindent Interestingly, for some cases like subject 105 running, even after giving 25 examples each for `running' and `walking', the LLM suspected an outlier sample and gave the following response.\\

\noindent
\textit{``Based on the provided data, it is not possible to accurately determine the activity (running or walking) directly, as the new reading [5.1111 1.7958 7.9847] falls outside the range of values specified for both running and walking. However, If the above data is the only information we have and we need to make a simplified assumption based on it,\textbf{ we might say the data could be closer to `walking' since the X and Y-axis readings are closer to the walking readings than the running ones}, although it is a high-level assumption and not very accurate due to the significant deviation on the Z-axis.''}

\subsection{Lessons Learnt}
One of the critical observations we get through this set of motivational experiments in Phase 1 is that although GPT-4 can understand the mapping of the values to the individual axes of the accelerometer, it cannot comprehend the accurate mapping between the input sample and the final label. Notably, we clearly observed that even with examples, the reasoning given by GPT-4 does not improve, and it lacks consistency in answering.
\section{Phase 2: Encoding-based In-context Learning}
\label{system}
\begin{figure*}
    \centering
    \includegraphics[width=\columnwidth,keepaspectratio]{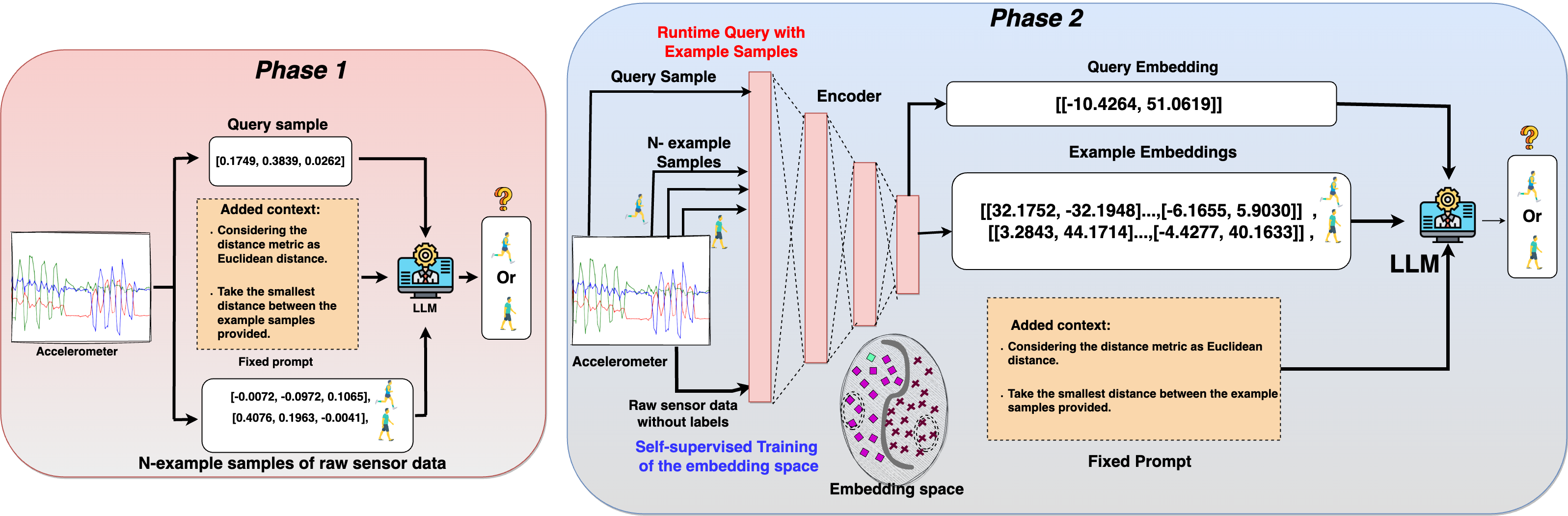}
    \caption{The broad overview of the setup that the designed study investigates. This shows an example scenario with data from two classes presented for the task of annotation.}
    \label{fig:overview_phase}
\end{figure*}
\textbf{Objective and Broad Intuition:} One of the critical concerns we saw in Phase 1 was the zero-improvement in the response generated by the LLMs, even with additional examples provided with the prompt. Considering this, we, in this phase, attempt to project the raw sensor into an embedding space such that the examples provide enough context information to the LLM through their neighborhood information. Naturally, we proceed in this direction by first including expert knowledge of computing the distance in the prompt and looking into different encoding strategies. A summary of the phases is provided in \figurename~\ref{fig:overview_phase}, and the details follow.
\subsection{Adding Context with Metric-based Guidance}
Recent works like~\cite{xu2023penetrative} have shown that one of the key ideas for obtaining accurate responses from the LLM can be by providing expert knowledge to the LLM through the prompt itself. As our broad intuition is to exploit the neighborhood of the embedding space as a critical information we specifically mention in the prompt to look for the distance between the example and the query sample (see Section~\ref{query_redesign}). This paper profoundly investigates this metric-based guidance with different standard distance metrics. 
\subsection{Encoding the Raw Sensor Data}
As mentioned above, one of the critical aspects of this encoding scheme is to project the raw sensor data into an embedding space, where we can exploit the neighborhood of the samples to retrieve the exact label given a query sample. Interestingly, we see that self-supervised contrastive learning allows us to achieve this even without access to the ground-truth labels~\cite{jaiswal2020survey,chen2020simple,spathis2022breaking,he2020momentum,yuan2022self}. Broadly, self-supervised contrastive learning algorithms on time-series data can be categorized into two broad categories -- (a) methods that look into the time-domain aspects of the signal~\cite{tang2020exploring,ramamurthy2022cogax}, and (b) methods that look beyond the time-domain into frequency and subsequence information as well~\cite{zhang2022self,xu2023retrieval}. More specifically, we investigate two algorithms from each of these categories and study their impact on the performance of LLMs in understanding the encoded data. The rationale and the details of the algorithms are summarised as follows.
\subsubsection{Time-domain Representations using Augmentations}
Classical self-supervised learning (SSL) algorithms on images~\cite{chen2020simple} are designed to learn visual representations without labeled data. It operates by creating multiple augmented views of the same image and uses a contrastive loss function to train a model to identify which view originates from the same image. Specifically, the contrastive loss function minimizes the distance between the embeddings of similar items while maximizing the distance between the embeddings of dissimilar ones. To ensure that the positive pairs stay closer than the negative pairs in the latent space, a typical contrastive algorithm optimizes the NT-Xcent loss~\cite{chen2020simple} defined as follows.
\[
\mathcal{L}_{\text{SimCLR}}(i, j) = -\log \frac{\exp(\text{sim}(z_i, z_j) / \tau)}{\sum_{k=1}^{2N} \mathbbm{1}_{[k \neq i]} \exp(\text{sim}(z_i, z_k) / \tau)}
\]
where the term $\text{sim}(u, v)$ represents the cosine similarity between two vectors $u$ and $v$. The parameter $\tau$ is the temperature parameter. The variable $N$ denotes the batch size, which effectively becomes $2N$ due to the augmentation strategy. Lastly, the expression $\mathbbm{1}_{[k \neq i]}$ is an indicator function that equals 1 if and only if $k$ is not equal to $i$. This approach is further extended by works like~\cite{tang2020exploring} for time-series inertial data, where they adopt a similar approach with appropriate augmentations like adding Gaussian Noise or scaling, which are more appropriate for the time-series-based data.
\subsubsection{Utilizing Frequency-domain Representations}
\begin{figure*}[]
    \centering
    \begin{minipage}{0.5\textwidth}
        \centering
        \includegraphics[width=\columnwidth,keepaspectratio]{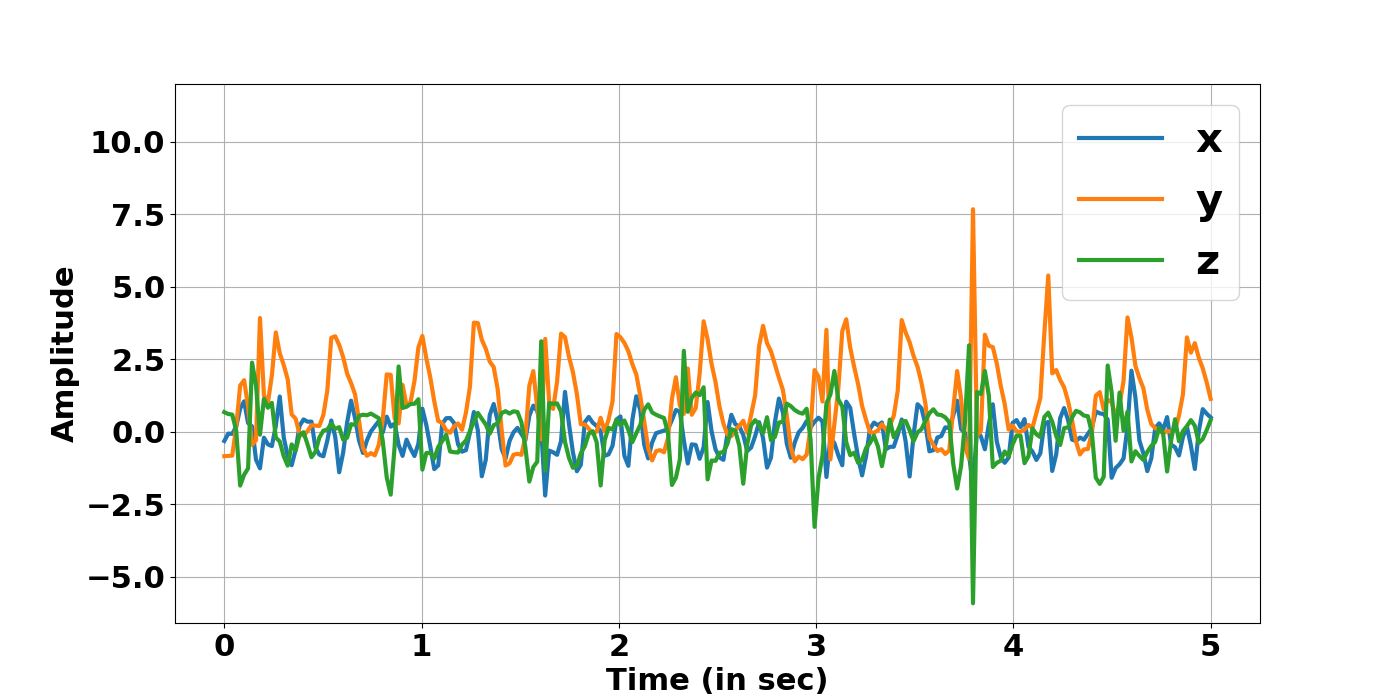}
        \subcaption{}
        \label{fig:time_rep_jogging}
    \end{minipage}%
    \begin{minipage}{0.5\textwidth}
        \centering
        \includegraphics[width=\columnwidth,keepaspectratio]{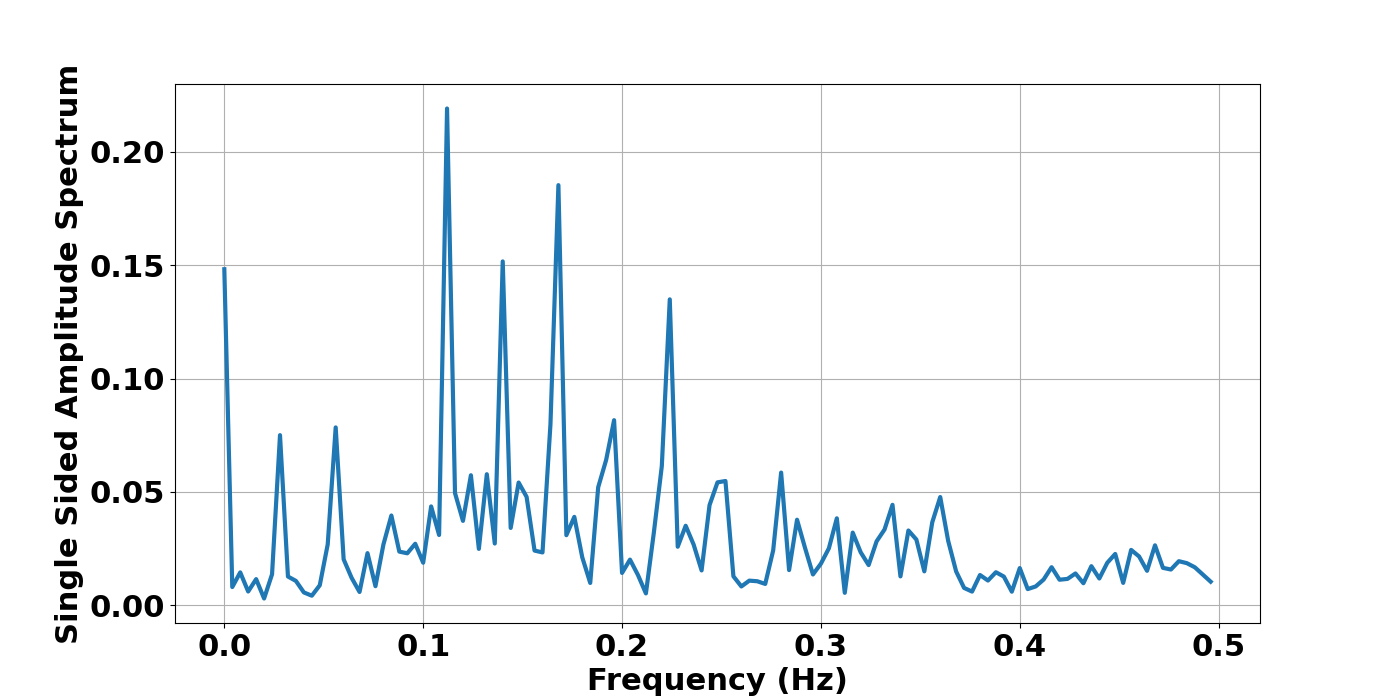}
        \subcaption{}
        \label{fig:freq_rep_jogging}
    \end{minipage}\\
    \begin{minipage}{0.5\textwidth}
        \centering
        \includegraphics[width=\columnwidth,keepaspectratio]{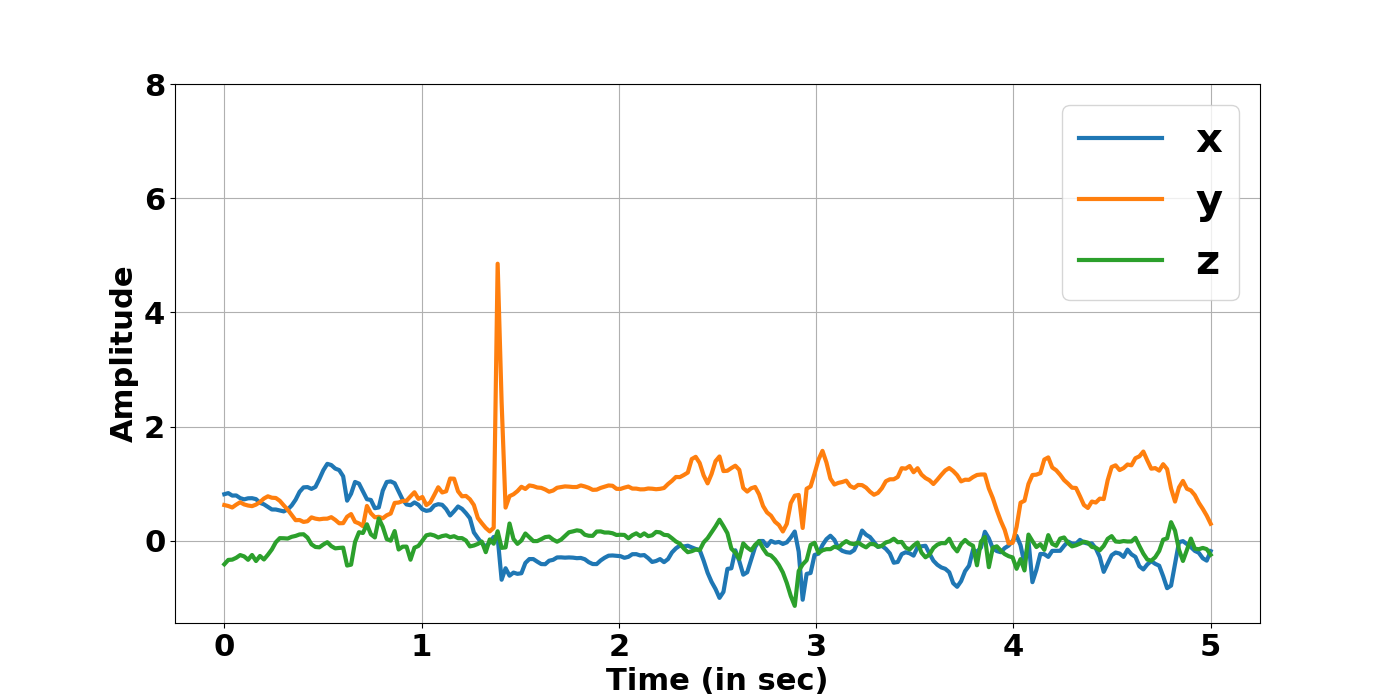}
        \subcaption{}
        \label{fig:time_rep_upstairs}
    \end{minipage}%
    \begin{minipage}{0.5\textwidth}
        \centering
        \includegraphics[width=\columnwidth,keepaspectratio]{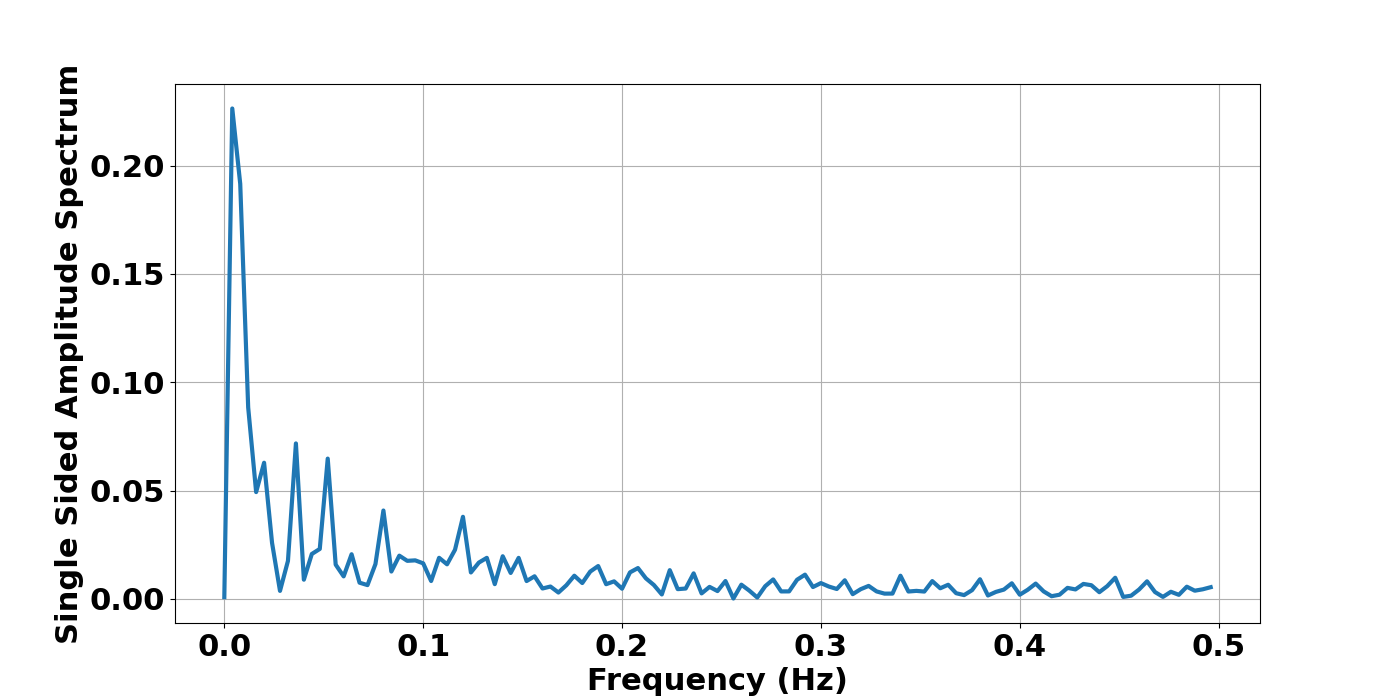}
        \subcaption{}
        \label{fig:freq_rep_upstairs}
    \end{minipage}
    \caption{Time and frequency domain analysis of a time-series data from MotionSense dataset across two different activity classes (a) and (b) correspond to time and frequency domain information for the activity ``Jogging'' whereas (c) and (d) correspond to time and frequency domain information for the activity ``Upstairs''.}
    \label{fig:time_frequency_domains_motionsense_dataset}
\end{figure*}

In addition to the general time-domain analysis, a deeper inspection of the frequency-domain information has often provided more insights, especially when the time-series data is noisy. Take, for example, the time and frequency-domain information for the two activities, ``\textit{Jogging}'' and ``\textit{Upstairs}'' from the MotionSense dataset~\cite{malekzadeh2019mobile}. We can observe from \figurename~\ref{fig:time_frequency_domains_motionsense_dataset} that the frequency-domain representations can better differentiate between the two activities. Understanding this, we next choose the \textit{Time-Frequency Contrastive} (TFC) learning approach~\cite{zhang2022self} that allows us to extract both frequency and time-domain domain embeddings for a given time series. The key behind this approach is that the time and frequency-based embeddings of a time-series sample should be close to each other in the latent time-frequency space, even when learned from the augmentations of the same sample. To achieve this, TFC uses two encoders, one for generating time-based embeddings (say, $G_{T}$ ) and the other for generating frequency-based embeddings (say, $G_{F}$). Later, the representations coming from these two encoders are mapped to a joint time-frequency space to measure the distance between the embeddings while optimizing the pre-training loss defined as follows.
The~\textbf{Time-based contrastive loss} is expressed as,
\[
\mathcal{L}_{T,i} = d(\mathbf{h}_i^T, \tilde{\mathbf{h}}_i^T, D) = -\log \frac{\exp(\text{sim}(\mathbf{h}_i^T, \tilde{\mathbf{h}}_i^T) / \tau)}{\sum_{x_j \in D} \mathbbm{1}_{[i \neq j]}\exp(\text{sim}(\mathbf{h}_i^T, G_T(x_j)) / \tau)}
\]

\noindent
The~\textbf{Frequency-based contrastive loss} is expressed as,
\[
\mathcal{L}_{F,i} = d(\mathbf{h}_i^F, \tilde{\mathbf{h}}_i^F, D) = -\log \frac{\exp(\text{sim}(\mathbf{h}_i^F, \tilde{\mathbf{h}}_i^F) / \tau)}{\sum_{x_j \in D} \mathbbm{1}_{[i \neq j]}\exp(\text{sim}(\mathbf{h}_i^F, G_F(x_j)) / \tau)}
\]

\noindent
Finally, the~\textbf{Time-Frequency contrastive loss} is expressed as,
\[
\mathcal{L}_{C,i} = \sum{S_{\text{pair}}} (\lvert S_i^{TF} - S_i^{\text{pair}} \rvert + \delta), \quad S^{\text{pair}} \in \{ S_i^{T\tilde{F}}, S_i^{\tilde{T}F}, S_i^{\tilde{T}\tilde{F}} \}
\]

\noindent
Overall, the pre-training loss is computed as,
\[
\mathcal{L}_{TFC,i} = \lambda(\mathcal{L}_{T,i} + \mathcal{L}_{F,i}) + (1 - \lambda)\mathcal{L}_{C,i}
\]

\noindent
where $\mathbf{h}i^T$ and $\mathbf{\tilde{h}}i^T$ denote the embeddings obtained from $G_{T}$, representing the transformation in the time domain. Similarly, $\mathbf{h}i^F$ and $\mathbf{\tilde{h}}i^F$ are the embeddings derived from $G_{F}$, which process the frequency domain aspects of the data. The dataset used for pre-training is denoted by ~\textit{D}. The function $\text{sim}(u, v)$ is utilized to compute the cosine similarity between vectors $u$ and $v$, facilitating a measure of their alignment. The parameter $\tau$ serves as the temperature coefficient, adjusting the distribution of similarity scores. The indicator function $\mathbbm{1}_{[k \neq i]}$ yields a value of 1 if and only if the indices $k$ and $i$ refer to different elements, ensuring the exclusion of self-comparisons. Each input time-series sample is represented by $x_j$. The distances between the transformed embeddings, such as $S^{TF}_i = d(z^{T}_i, z^{F}_i, D)$ and its variations including $S_i^{T\tilde{F}}$, $S_i^{\tilde{T}F}$, and $S_i^{\tilde{T}\tilde{F}}$, encapsulate the relative positions of embeddings in the transformed space. Here, $z_i^T = R_T(G_T(x_i^T))$ and $z_i^F = R_F(G_F(x_i^F))$ are the results of applying cross projectors $R_T$ and $R_F$ respectively, ensuring that the embeddings from $G_T$ and $G_F$ are aligned within a joint time-frequency space.

\subsection{Annotation as Retrieval of Labels using Examples}
\begin{figure*}[]
    \centering
    \begin{minipage}{0.5\textwidth}
        \centering
        \includegraphics[width=\columnwidth,keepaspectratio]{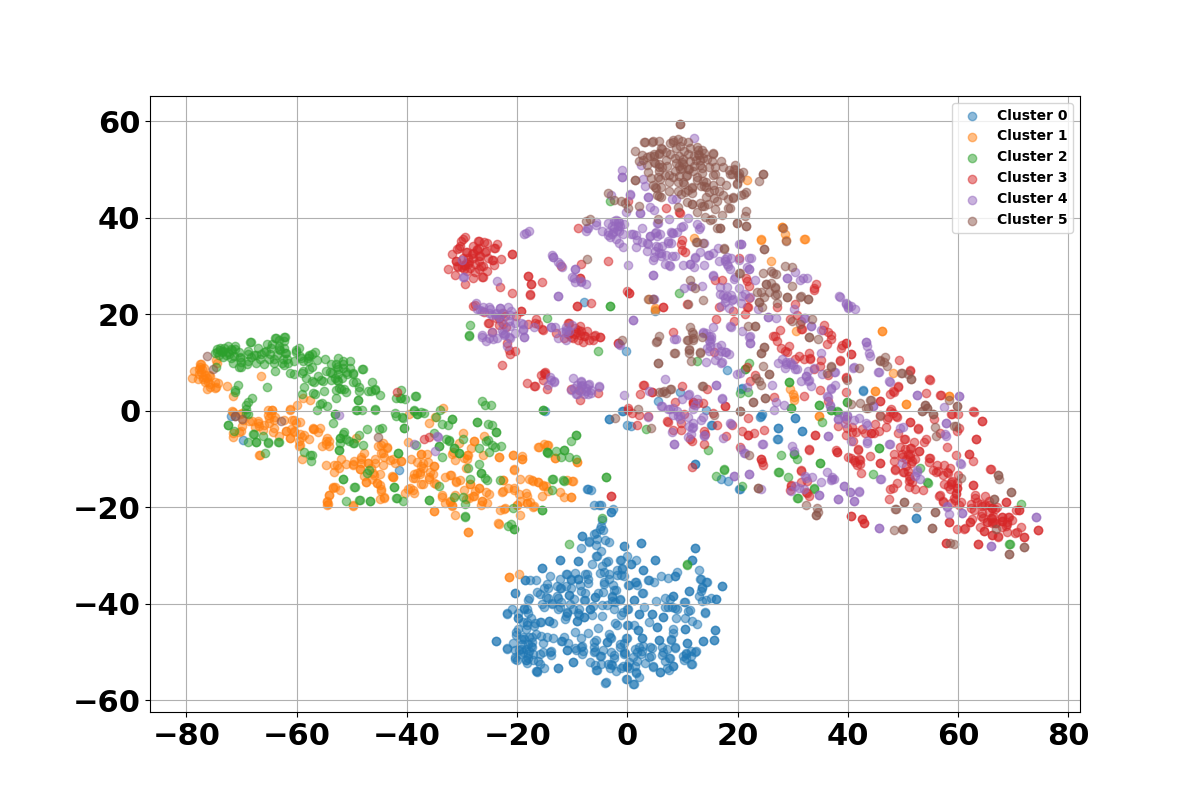}
        \subcaption{}
        \label{fig:simclr_domain_embeds}
    \end{minipage}%
    \begin{minipage}{0.5\textwidth}
        \centering
        \includegraphics[width=\columnwidth,keepaspectratio]{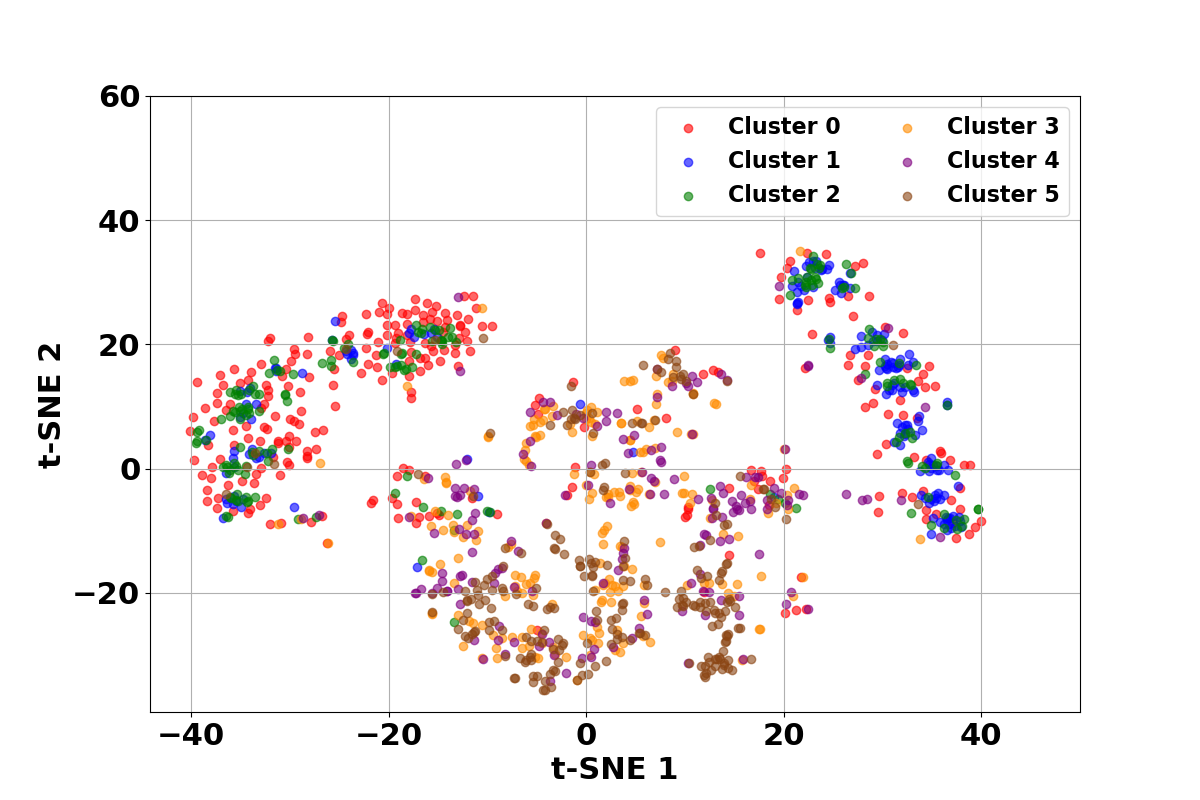}
        \subcaption{}
        \label{fig:time_domain_embeds}
    \end{minipage}\\
    \begin{minipage}{0.5\textwidth}
        \centering
        \includegraphics[width=\columnwidth,keepaspectratio]{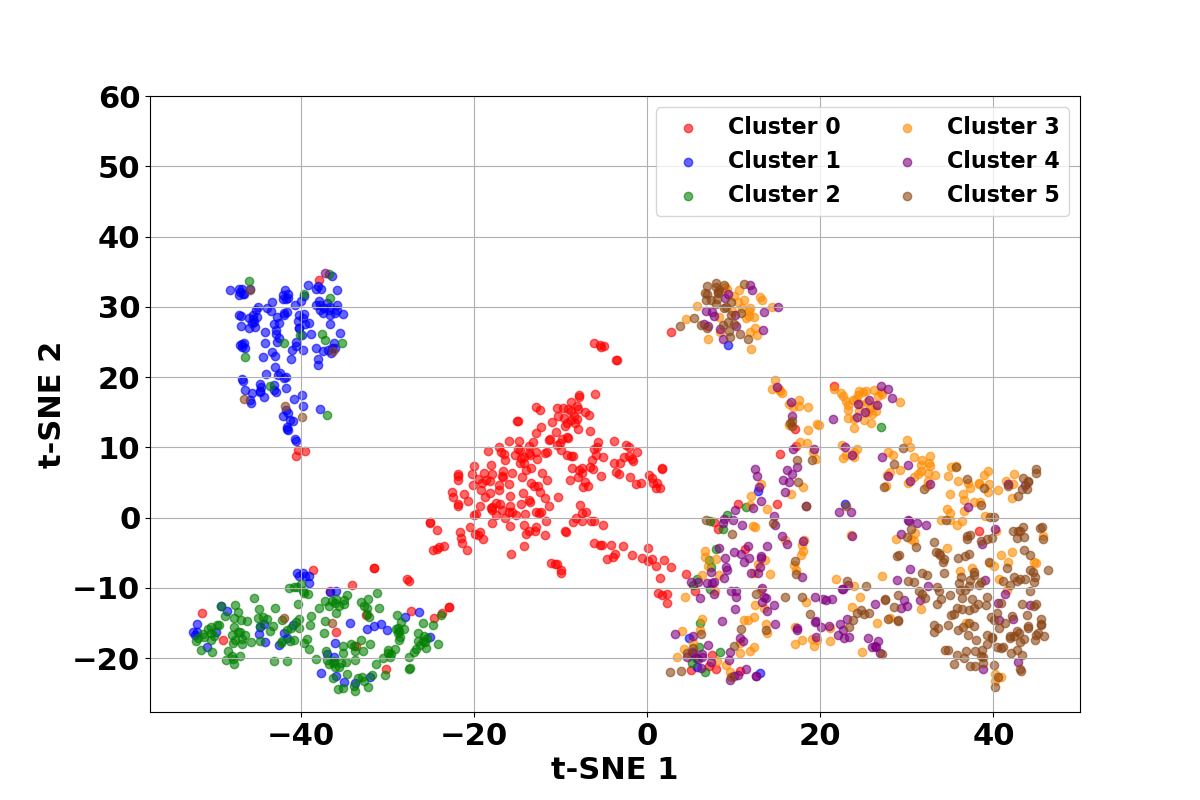}
        \subcaption{}
        \label{fig:freq_domain_embeds}
    \end{minipage}%
    \begin{minipage}{0.5\textwidth}
        \centering
        \includegraphics[width=\columnwidth,keepaspectratio]{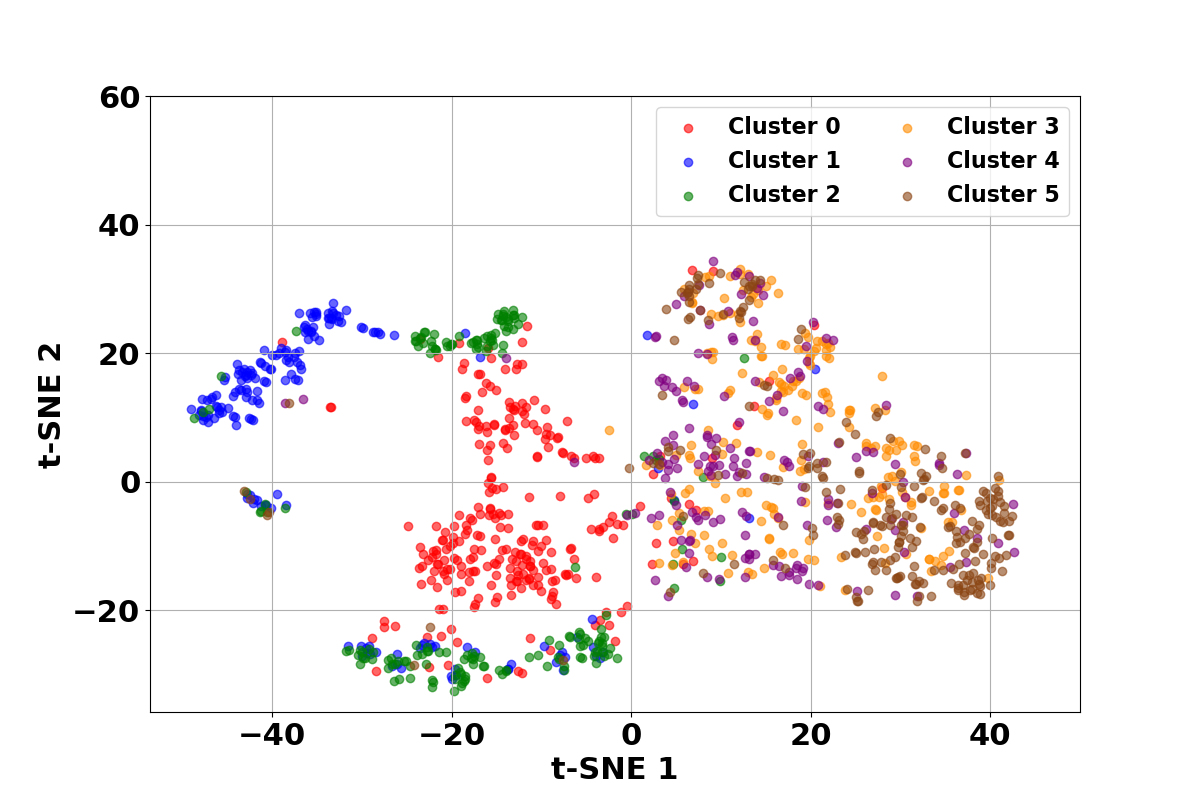}
        \subcaption{}
        \label{fig:all_domain_embeds}
    \end{minipage}%
    \caption{t-SNE plots of (a) SimCLR and (b,c,d) TFC embeddings for the HHAR dataset~\cite{misc_heterogeneity_activity_recognition_344}. Here, (b) time domain, (c) frequency domain, and (d) both time and frequency domain embeddings concatenated obtained from TFC-based encoder.}
    \label{fig:embeds}
\end{figure*}

The trained encoders allow the raw sensor data to be projected to the embedding space where the samples from the same class cluster in space (see \figurename~\ref{fig:embeds}). Also, we observe that for the TFC-based encoder, the aggregated embeddings from both the time and frequency domains provide more distinct clusters in the embedding space in comparison to the individual domain-based representations. Nevertheless, once the encoders are trained, the next task of obtaining annotation labels from the LLM becomes analogous to a query-based retrieval, given some examples drawn from the embedding space. \textbf{The objective of this study is to assess whether this approach of providing examples from the embedding space provides enough contextual information for the LLM to provide accurate and consistent annotations for a query sample with proper reasoning.} The details of the assessment and the primary observations are summarized in the next section. 

\section{Empirical Results and Analysis}
\label{eval}
In the remainder of this paper, we now investigate the performance benefits of using the SSL-based encoded sensor data in place of raw sensor data and lay out a few strategies for redesigning the prompt to incorporate the embeddings as a part of the query. The basic setup, design choices, key observations, and the final takeaways are summarized below.
\subsection{Implementation and Hyperparameters}
\subsubsection{SimCLR}
In this implementation of SimCLR on time-series data, SGD optimizer with a cosine decay of learning rate 0.1 is used for pre-training for 200 epochs having a mini-batch size of 512 for MotionSense, PAMAP2, and HHAR datasets, respectively. However, for the UCI HAR dataset, we have taken the learning rate as 0.001 and the mini-batch size as 64. For the implementation, we have adapted the official implementation of the paper~\cite{tang2020exploring} available here\footnote{\url{https://github.com/iantangc/ContrastiveLearningHAR}}.
\subsubsection{TFC}
For training, Adam optimizer ~\cite{kingma2014adam} is used with $\beta_1 = 0.9$ and $\beta_2 = 0.99$. The learning rate used is $3e-4$. The mini-batch size is 128. The training epochs are 40. The above hyperparameters are used for all the four datasets we have considered. For implementing the TFC-based contrastive learning, we followed the official implementation available here\footnote{\url{https://github.com/mims-harvard/TFC-pretraining}}.
\subsection{Redesigning the Prompt}
\label{query_redesign}
Although this paper is not focused on performing sophisticated prompt-tuning, we needed to change the initial set of queries we designed for the motivational experiment (See Section~\ref{query_motiv}). These changes were made to accommodate the replacement of embeddings in place of raw sensor readings. Furthermore, we introduce newer contextual information by adding information regarding the distance metrics to enable the LLM to find better reasoning in terms of similarity (or proximity) between the query and the example samples. More specifically, the redesigned prompt is designed as follows.\\

\noindent
\textit{The following given embeddings correspond to ``class 1'':  $[[\mathbb{E}^1_{11},\hdots,\mathbb{E}^1_{u1}]\hdots[\mathbb{E}^1_{1v},\hdots,\mathbb{E}^1_{uv}]]$, the following given embeddings correspond to ``class 2'': $[[\mathbb{E}^2_{11},\hdots,\mathbb{E}^2_{u1}]\hdots[\mathbb{E}^2_{1v},\hdots,\mathbb{E}^2_{uv}]]$ . . . . . . and the following given embeddings corresponds to ``class n'': $[[\mathbb{E}^n_{11},\hdots,\mathbb{E}^n_{u1}]\hdots[\mathbb{E}^n_{1v},\hdots,\mathbb{E}^n_{uv}]]$ classify the embedding $[\mathbb{E}^q_{11},\hdots,\mathbb{E}^q_{1u}]$ as either ``class 1'' or ``class 2'' . . . . . . . or ``class n'' considering the minimum distance to the example embeddings provided that the distance metric chosen is euclidean distance.}\\

\noindent
Here $u$ and $v$ represent the dimension and total number of example embeddings, respectively. In this study, We change both $u$ and $v$ to study the impact of the dimension of embeddings. Furthermore, we investigate the impact of distance metrics in assisting the LLM in arriving at a conclusion regarding the final response.

\subsection{Designing the Experiments}
Although a straightforward analysis can easily be done just by comparing the accuracy of the responses obtained from GPT-4 with the ground truth, there are subsequently different implicit factors like token length and cost of using this approach of annotation that may finally impact the potential of using LLM as virtual sensors. Thus, to perform this study elaborately, we redesign the set of experiments to investigate the impact of different factors on accuracy while also considering their impact on the length of the final query, cost, and time required for annotation.
\subsubsection{Studying the Impact of Dimensions} To study the impact of the dimensions, we first fix the choices of distance and number of examples as Euclidean and one, respectively. Next, we vary the dimensions from 2 to 15 and record the changes in the accuracy of the response. To reduce the original embedding dimension (96 in the case of SimCLR and 668 for TFC), we use t-SNE as the unsupervised dimensionality reduction approach. Although a higher accuracy is a crucial factor, \textbf{an increase in the dimension will actually lead to a significant increase in the number of tokens} thus leading to \textbf{higher cost of annotation}~\cite{gpt_cost}.
\subsubsection{Studying Impact of Distance} The purpose of including the distance metric is to provide additional reference to the LLM to assess the similarity (or the proximity in the embedding space) between the queried and the example samples. To study the impact of the distance metric, we first fix the number of examples as 3 and the embedding dimension as 2, except for the baseline case with no encoding of the input where we keep the dimension as 3 (input from the triaxial accelerometer). We then vary the distance metrics between Euclidean, Manhattan, and Cosine. Here, the \textbf{key factor of analysis is accuracy} only as the \textbf{change in the distance metric type does not impact the number of tokens significantly}.
\subsubsection{Analysing the Impact of Number of Examples} We study the impact of a number of examples by first fixing the distance metric as Euclidean and embedding dimension as 2. We then vary the number of examples to 1, 3, 10, and 25. Considering the analogy with a human-in-the-loop setup of\textbf{ a good annotation framework} will be a \textbf{higher accuracy with more examples}. However, \textbf{an increase in the number of examples} will also have a \textbf{significant increase in the number of tokens} and thus \textbf{will have a higher cost of annotation} with paid APIs like GPT-4~\cite{gpt_cost}.
\begin{figure*}[]
    \centering
    \begin{minipage}{0.48\textwidth}
        \centering
        \includegraphics[width=\columnwidth,keepaspectratio]{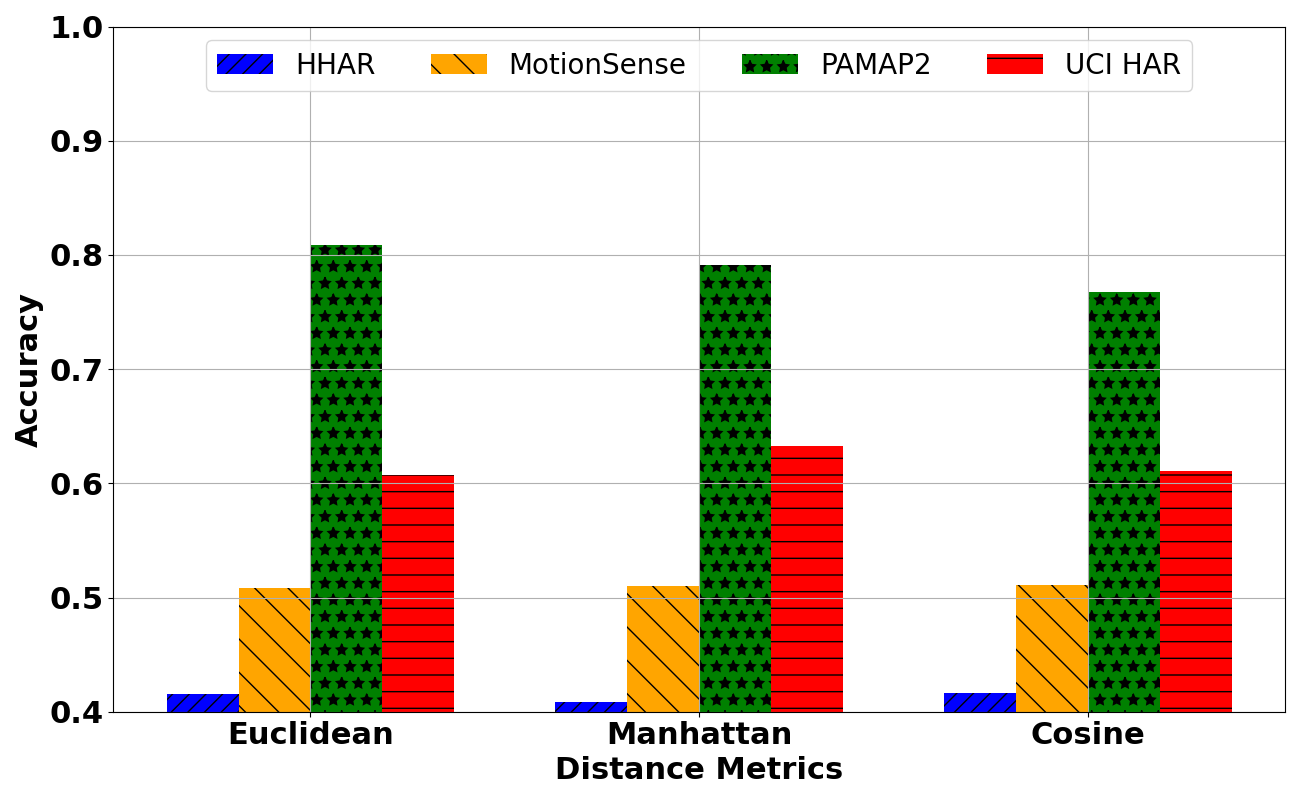}
        \subcaption{}
        \label{fig:baseline_distance}
    \end{minipage}%
    \begin{minipage}{0.48\textwidth}
        \centering
        \includegraphics[width=\columnwidth,keepaspectratio]{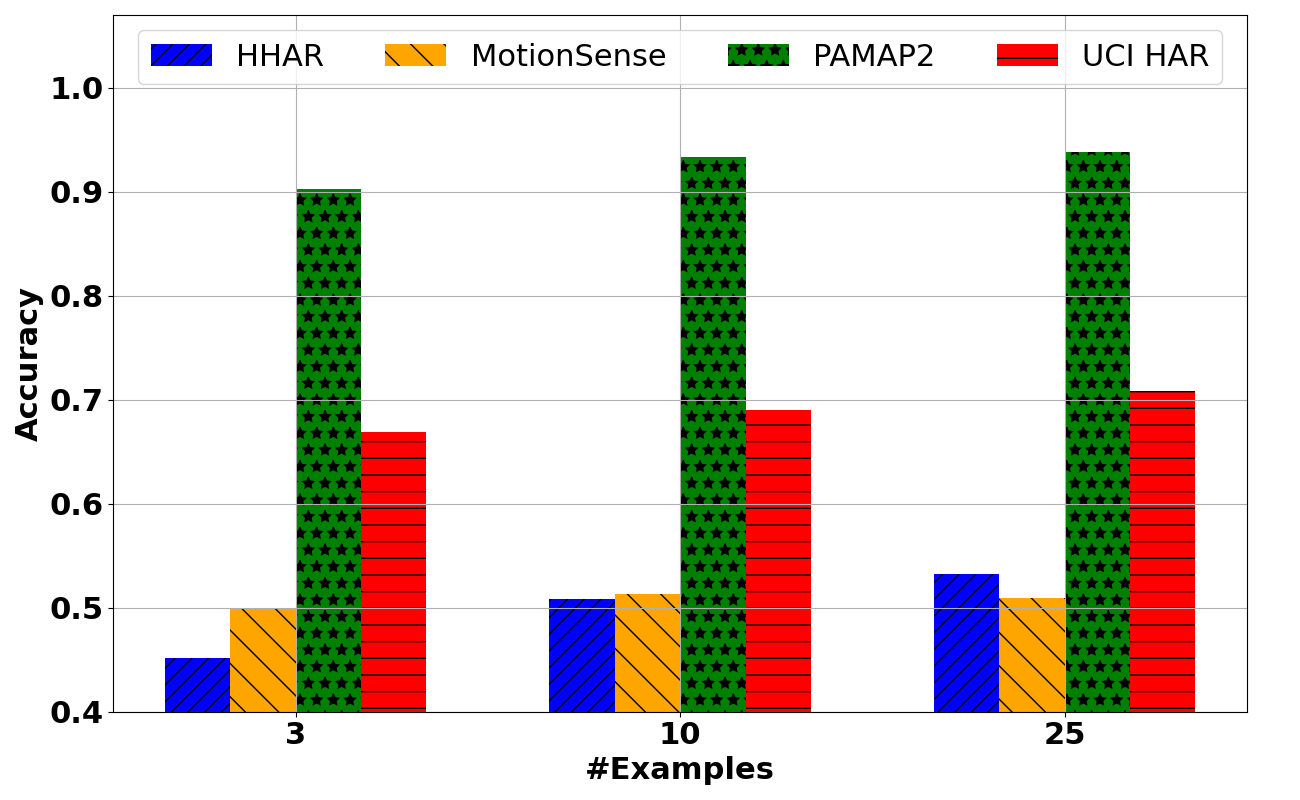}
        \subcaption{}
        \label{fig:baseline_shot}
    \end{minipage}
    \caption{Baseline accuracy of responses received from GPT-4 using raw sensor data across different HAR datasets considering -- (a) different distance metrics and (b) with different number of examples.}
    \label{fig:acc_base}
\end{figure*}
\begin{figure*}[]
    \centering
    \begin{minipage}{0.33\textwidth}
        \centering
        \includegraphics[width=\columnwidth,keepaspectratio]{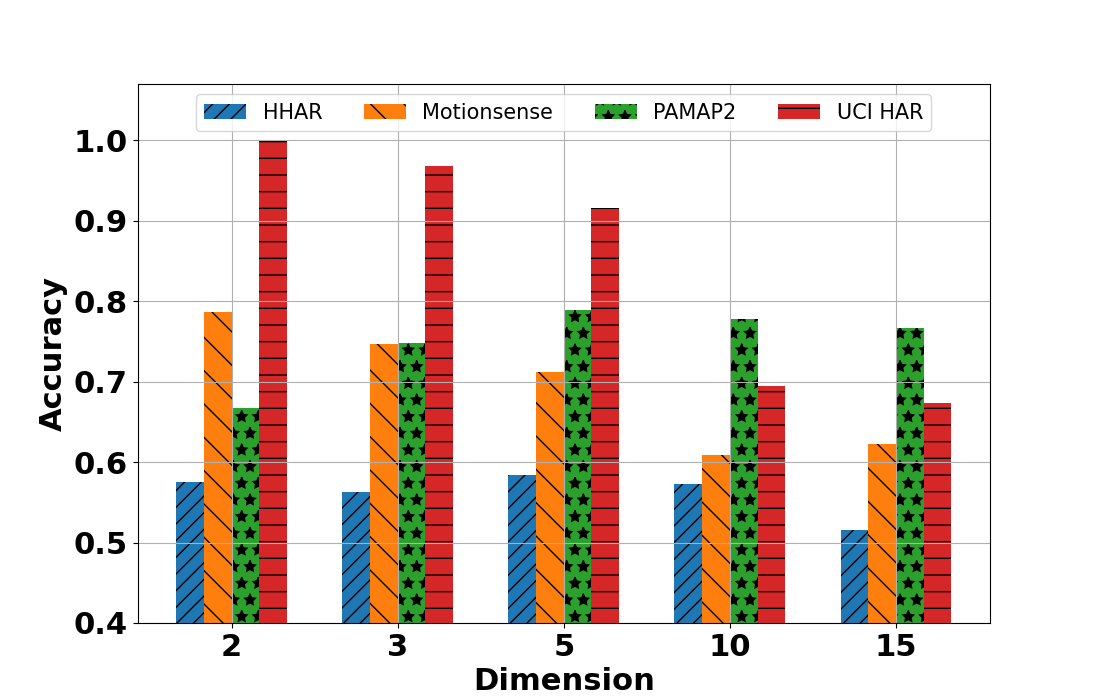}
        \subcaption{}
        \label{fig:sim_clr_dim}
    \end{minipage}%
    \begin{minipage}{0.33\textwidth}
        \centering
        \includegraphics[width=\columnwidth,keepaspectratio]{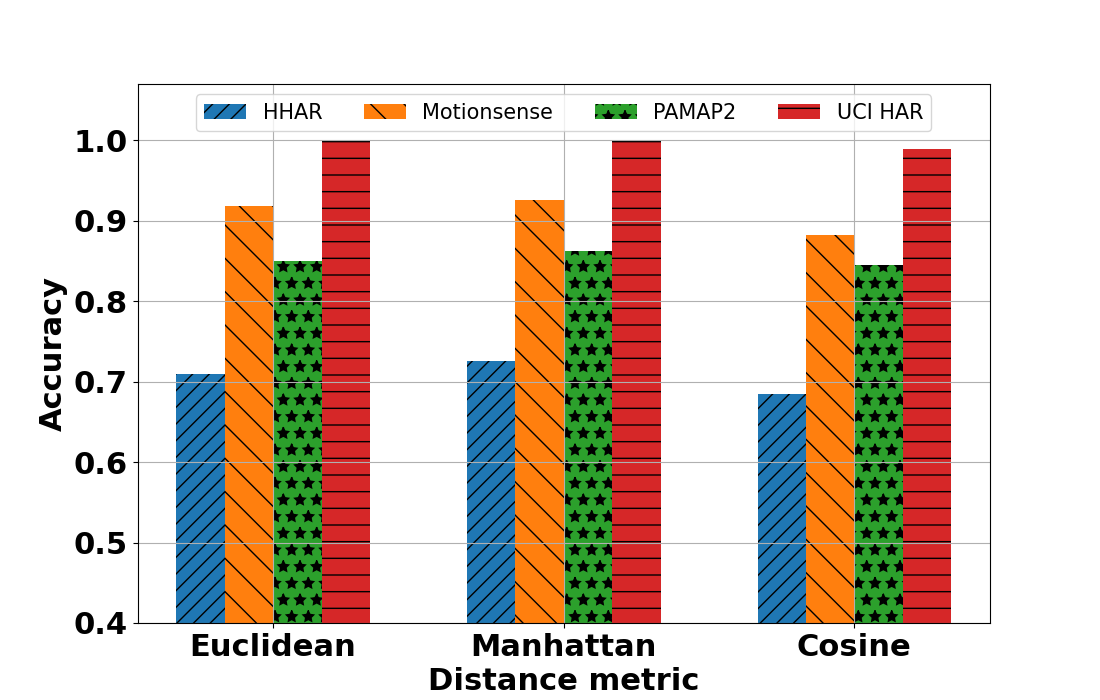}
        \subcaption{}
        \label{fig:sim_clr_dist}
    \end{minipage}%
    \begin{minipage}{0.33\textwidth}
        \centering
        \includegraphics[width=\columnwidth,keepaspectratio]{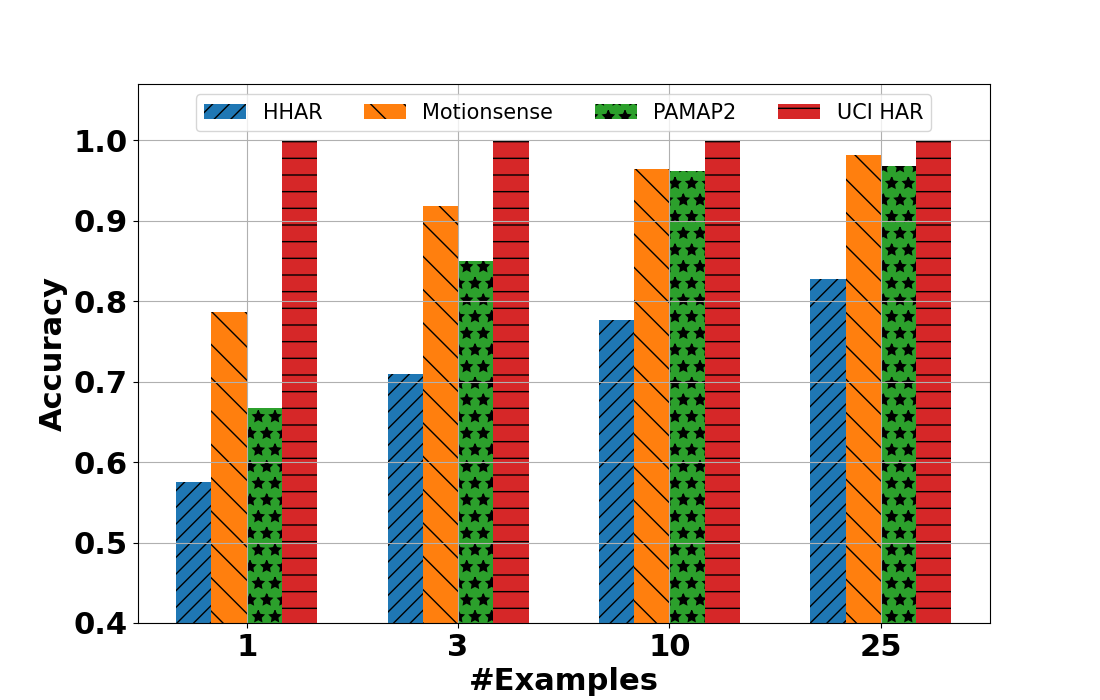}
        \subcaption{}
        \label{fig:sim_clr_shot}
    \end{minipage}
    \caption{Accuracy of the responses obtained from GPT-4 using the encoded output from a pre-trained encoder trained using the SimCLR approach. Here, we show the results by varying -- (a) the dimensionality of the embeddings, (b) the distance metric for comparing the embedding with the example embeddings included as a part of the query, and (c) the number of examples.}
    \label{fig:acc_simclr}
\end{figure*}

\subsection{Performance Accuracy}
\begin{figure*}[]
    \centering
    \begin{minipage}{0.33\textwidth}
        \centering
        \includegraphics[width=\columnwidth,keepaspectratio]{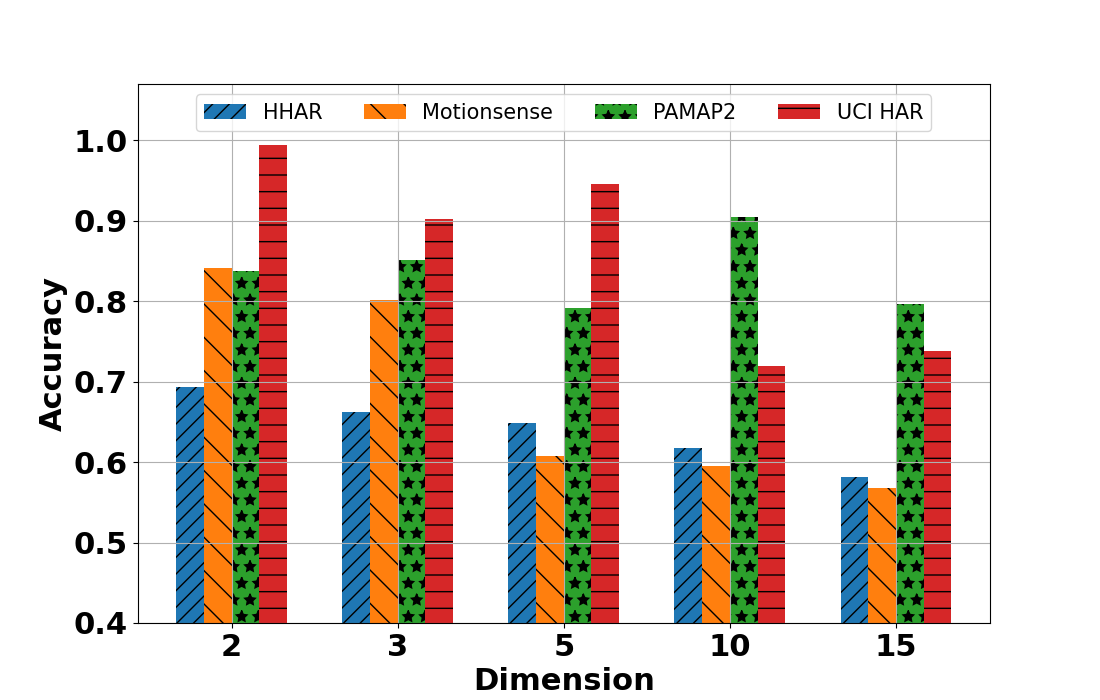}
        \subcaption{}
        \label{fig:tfc_dim}
    \end{minipage}%
    \begin{minipage}{0.33\textwidth}
        \centering
        \includegraphics[width=\columnwidth,keepaspectratio]{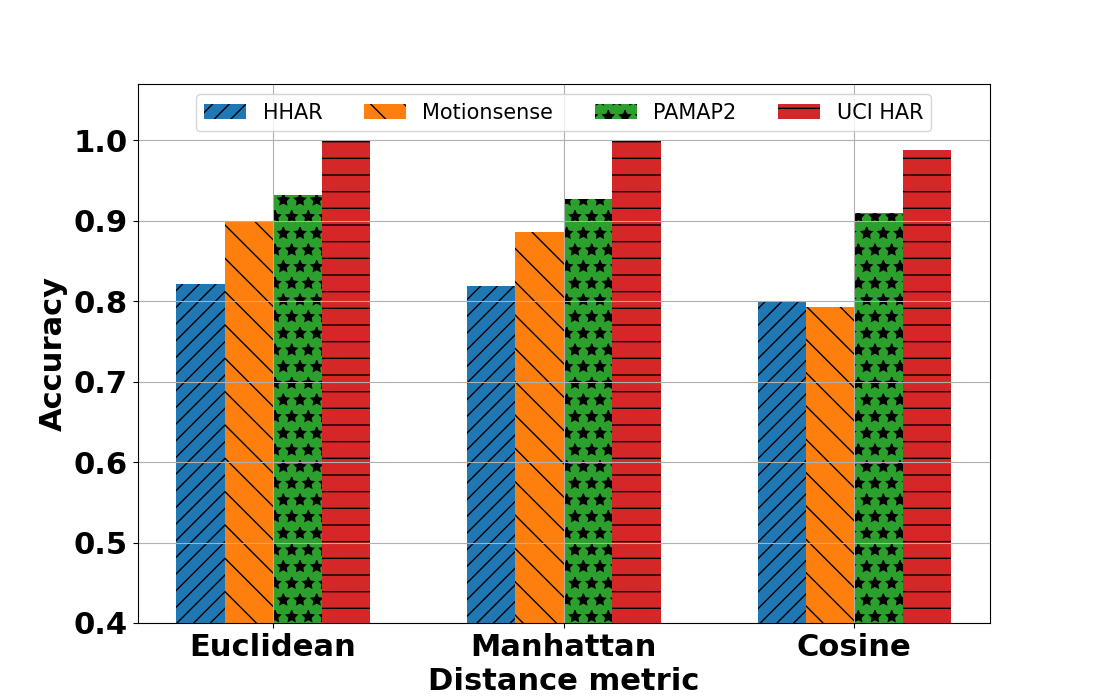}
        \subcaption{}
        \label{fig:tfc_dist}
    \end{minipage}%
    \begin{minipage}{0.33\textwidth}
        \centering
        \includegraphics[width=\columnwidth,keepaspectratio]{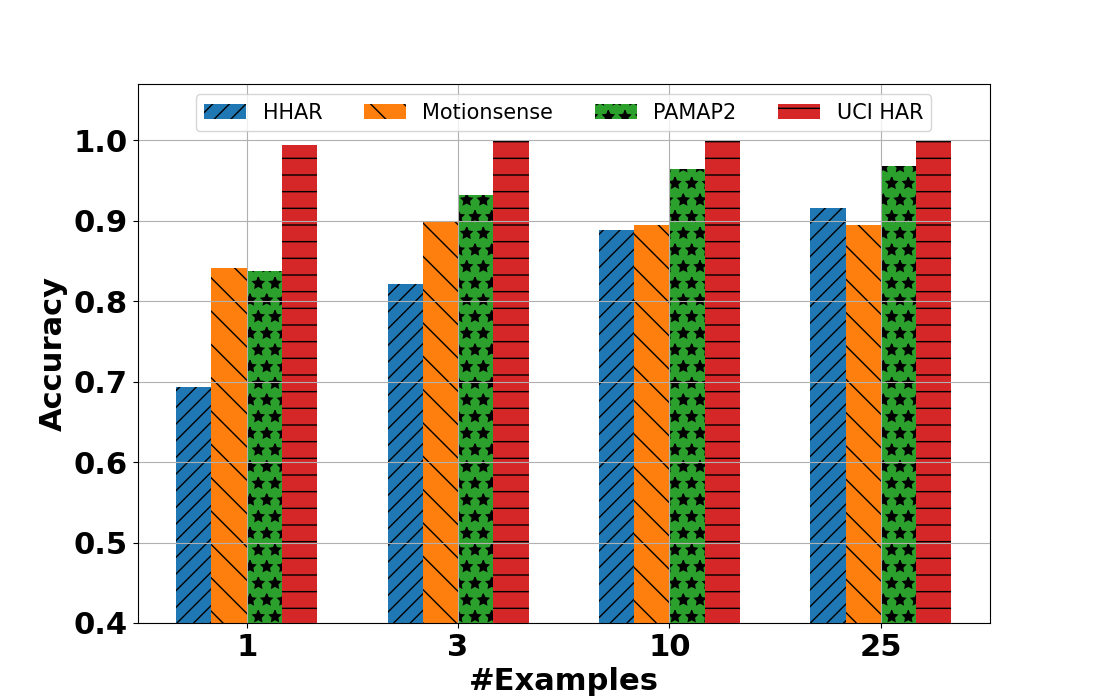}
        \subcaption{}
        \label{fig:tfc_shot}
    \end{minipage}
    \caption{Accuracy of the responses obtained from GPT-4 using the encoded output from a pre-trained encoder trained using the TFC approach. Here, we show the results by varying -- (a) the dimensionality of the embeddings, (b) the distance metric for comparing the embedding with the example embeddings included as a part of the query, and (c) the number of examples.}
    \label{fig:acc_tfc}
\end{figure*}

We start the analysis by first comparing the accuracy of annotations concerning factors like dimensionality of the embeddings, number of examples, and impact of the distance metric. For the results, we compare the baseline approach, with raw sensor data without any encoding presented in \figurename~\ref{fig:acc_base}, with the two encoding approaches of SimCLR and TFC presented in \figurename~\ref{fig:acc_simclr} and \figurename~\ref{fig:acc_tfc}, respectively. The summary of the observations is as follows.
\begin{figure}
    \centering
    \begin{subfigure}[b]{0.47\textwidth}
        \centering
        \includegraphics[width=\textwidth]{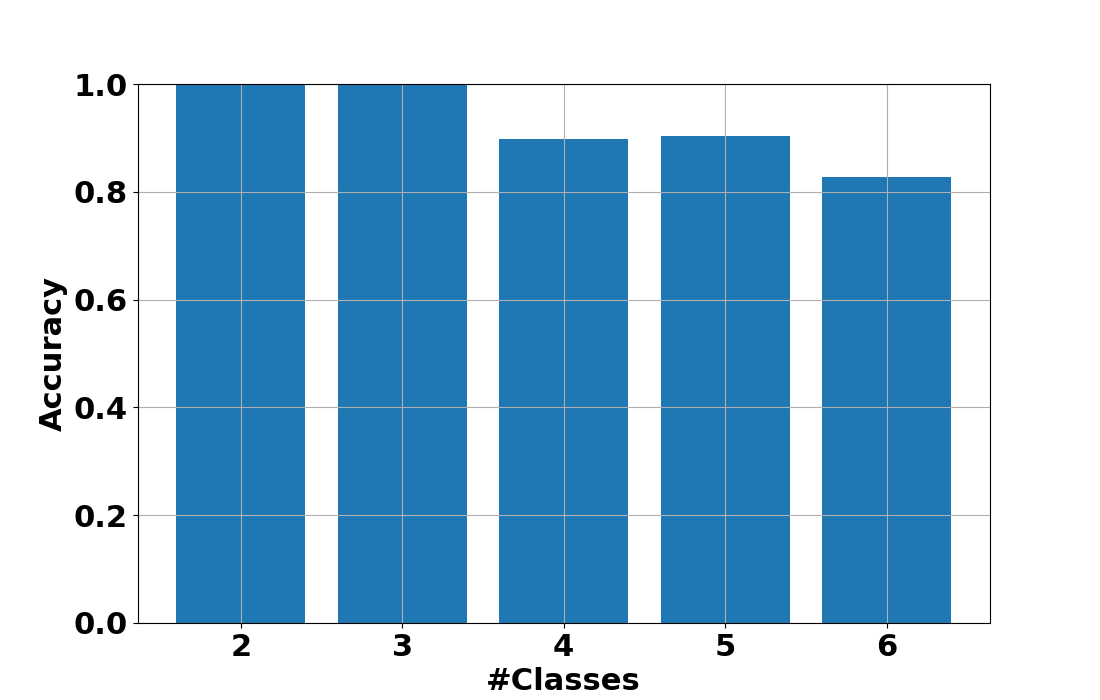}
        \caption{}
        \label{fig:figure1}
    \end{subfigure}%
    \begin{subfigure}[b]{0.47\textwidth}
        \centering
        \includegraphics[width=\textwidth]{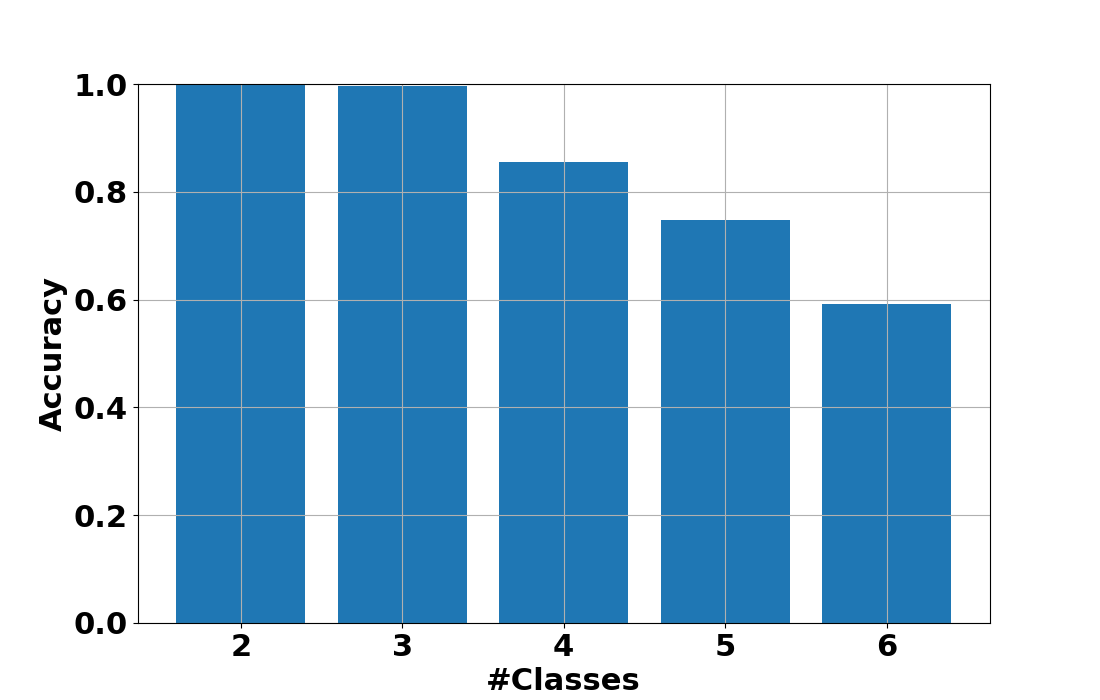}
        \caption{}
        \label{fig:figure2}
    \end{subfigure}
    \caption{Accuracy of annotations with increasing number of classes with embeddings generated using -- (a) SimCLR and (b) TFC. These results show that the accuracy drops with increasing number of classes irrespective of the contrastive setup used.}
    \label{fig:both_figures}
\end{figure}
\begin{figure}
  \centering
  \begin{subfigure}[b]{0.30\linewidth}
    \includegraphics[width=\linewidth]{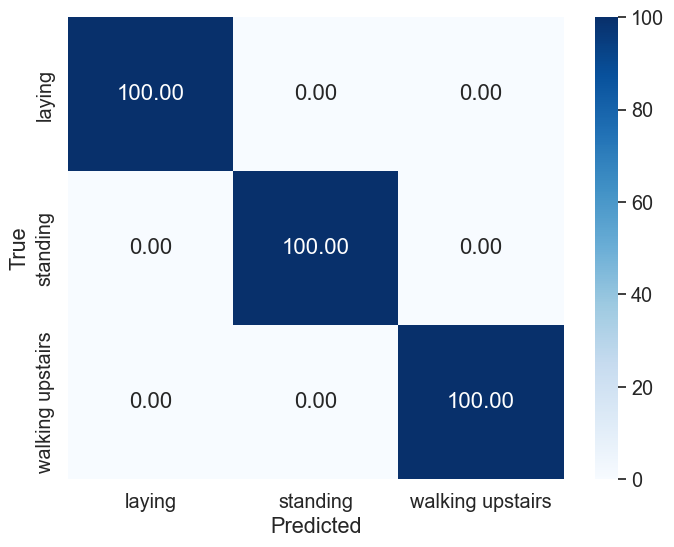}
    \caption{Three classes}
    \label{fig:a}
  \end{subfigure}
  \begin{subfigure}[b]{0.30\linewidth}
    \includegraphics[width=\linewidth]{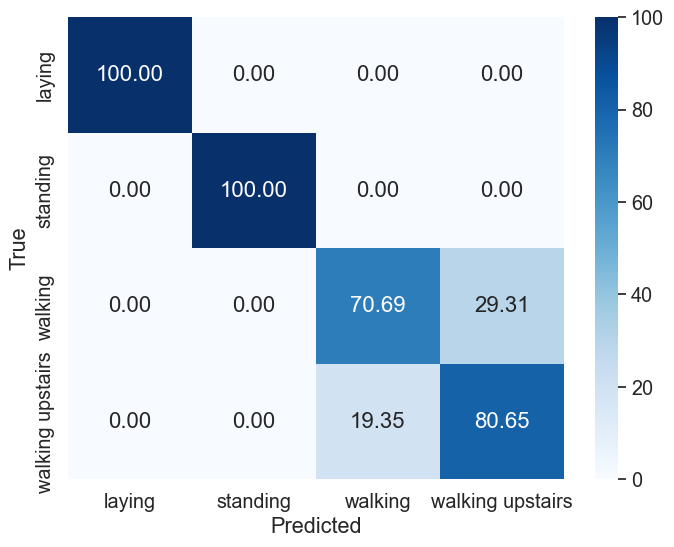}
    \caption{Four classes}
    \label{fig:b}
  \end{subfigure}
  
  \begin{subfigure}[b]{0.40\linewidth}
    \includegraphics[width=\linewidth]{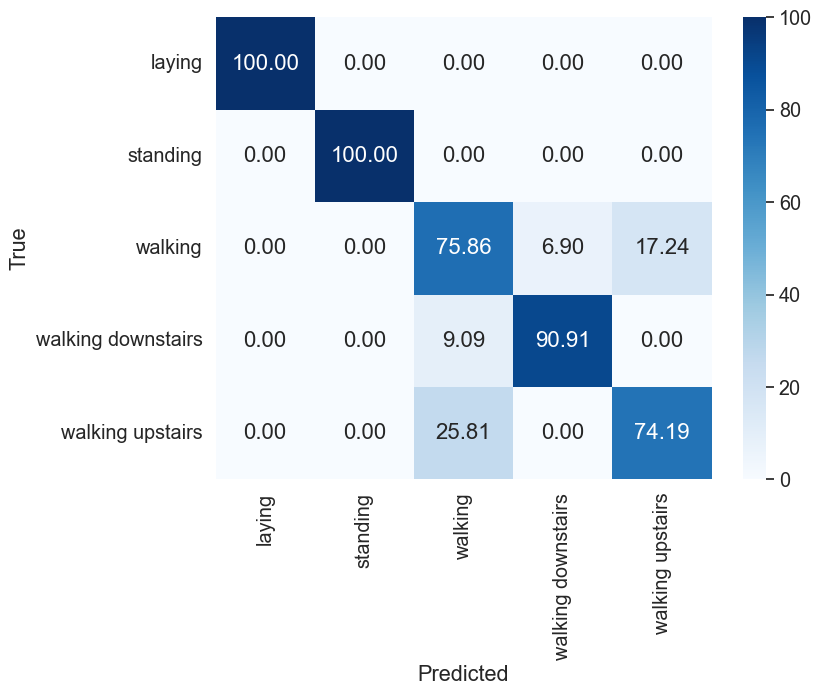}
    \caption{Five classes}
    \label{fig:c}
  \end{subfigure}%
  \begin{subfigure}[b]{0.40\linewidth}
    \includegraphics[width=\linewidth]{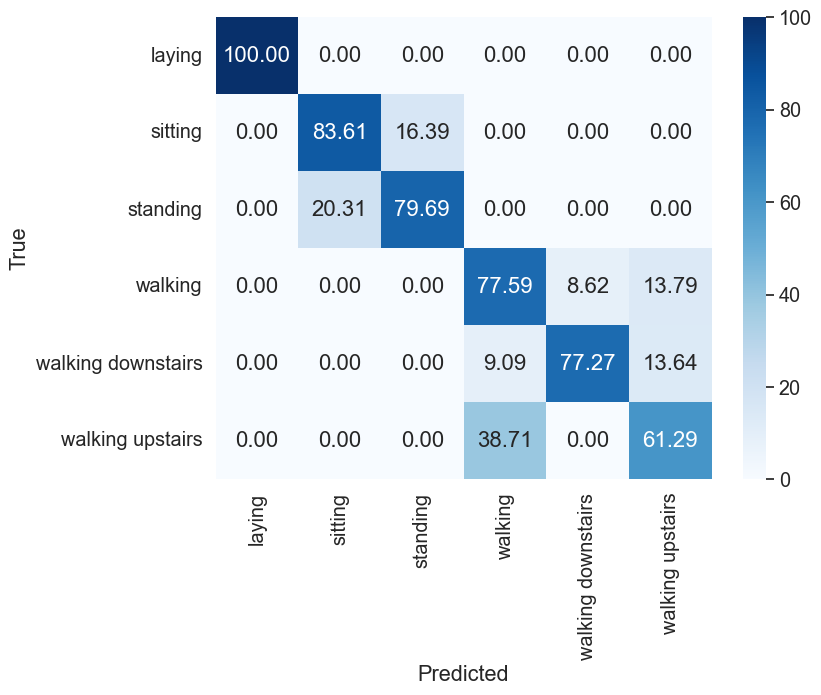}
    \caption{Six classes}
    \label{fig:d}
  \end{subfigure}
  \caption{Confusion matrices for multi-class scenarios with embeddings generated using SimCLR. These results expose the inherent limitation of using triaxial accelerometer data for differentiating closely related classes. For example, we see a high confusion from the annotations provided by the LLM with embeddings from closely related classes like ``walking'' and ``walking upstairs''.}
  \label{fig:conf_mat}
\end{figure}
\subsubsection{Impact of Dimension}
For this experiment, we do not consider the baseline approach as the raw sensor data from a triaxial accelerometer has a fixed dimension. For the encoded inputs through SimCLR (see \figurename~\ref{fig:sim_clr_dim}) and TFC (see \figurename~\ref{fig:tfc_dim}), we observe a similar pattern where the best accuracy is achieved at the reduced dimension of 2. Notably, for both these encoding approaches, we see that with higher dimensions, the accuracy of the responses drops significantly.
\subsubsection{Impact of Distance Metric}
We evaluate the accuracy of the responses across all datasets considering three distance metrics of Euclidean, Mahattan, and Cosine to evaluate the similarity of the queried sample with the examples given in the query. For all the cases, including the baseline case of using raw sensor data we do not observe a significant difference in the accuracy, albeit there is a slightly higher gain with Manhattan distance which is already known to be highly robust for high dimensional data (see \figurename~\ref{fig:baseline_distance}, \figurename~\ref{fig:sim_clr_dist}, and \figurename~\ref{fig:tfc_dist}).
\subsubsection{Impact of Number of Examples}
One key downside of using raw sensor data directly with LLMs that we observed in the motivational study and across all datasets in \figurename~\ref{fig:baseline_shot} is the lack of consistency with an increasing number of examples. However, with encoded (for both SimCLR and TFC) examples, we can clearly observe (from \figurename~\ref{fig:sim_clr_shot} and \figurename~\ref{fig:tfc_shot}) that the system achieves a higher consistency at times even reaching an accuracy of 1.0 when the number of examples is 25 (an average of 0.02\% ), the highest that we explore in this study. 

\begin{table}[]
\scriptsize
\centering
\caption{Impact of including distance metric as a context information in the query. The responses clearly show how GPT-4 then uses the distance metric as the basis for reasoning regarding the generated response.}
\label{tbl:reason_table}
\begin{tabular}{l|l|l}
\hline
\multirow{3}{*}{\begin{tabular}[c]{@{}l@{}}Raw\\ Sensor\\ Data\end{tabular}} &
  Query &
  \begin{tabular}[c]{@{}l@{}}\begin{minipage}[t]{0.8\columnwidth}The following given triaxial accelerometer readings correspond to ``jogging'': {[}{[}0.0223   0.1115   0.3571{]}...{[}0.161   0.1497   -0.2369{]}{]} and the following given triaxial accelerometer readings correspond to ``walk upstairs'': {[}{[}0.2615   0.118   -0.0637{]}...{[}-0.0322   0.0437   0.057{]}{]}; classify the triaxial accelerometer reading {[}-0.0933  0.0165  0.0609{]} as either "jogging" or "walking upstairs" provided that this data is coming from the smartphone stored in the user's trousers' front pocket considering the minimum distance to the example triaxial accelerometer readings provided that the distance metric chosen is euclidean distance. Answer in one word.\end{minipage}\tabularnewline\tabularnewline\end{tabular} \\ \cline{2-3} 
 &
  Response &
  \begin{tabular}[c]{@{}l@{}}
  \begin{minipage}[t]{0.8\columnwidth}The triaxial accelerometer reading {[}− 0.0933, 0.0165, 0.0609 {]} {[}−0.0933, 0.0165, 0.0609{]} is classified as ``walk upstairs'' based on the minimum Euclidean distance to the example triaxial accelerometer readings provided.\end{minipage}\tabularnewline\tabularnewline\end{tabular} \\ \cline{2-3} 
 &
  Reasoning &
  \begin{tabular}[c]{@{}l@{}} \begin{minipage}[t]{0.8\columnwidth} The classification of the triaxial accelerometer reading [ − 0.0933 , 0.0165 , 0.0609 {]} {[}−0.0933,0.0165,0.0609{]} as walk upstairs" is based on the principle of the nearest neighbor algorithm, using the Euclidean distance as the metric for similarity. This distance metric was chosen to compare the given accelerometer reading with each of the example readings from both activities ("jogging" and "walking upstairs"). Minimum Distance: For each activity, I calculated the Euclidean distance from the given reading to all example readings of that activity. Then, I found the minimum distance to each activity, determining the closest match for the given reading. Classification: The activity whose examples have the minimum distance to the given reading is considered the most similar or the most likely classification for that reading. In this case, the minimum distance between the given reading and the examples of "walk upstairs" was smaller than that to the examples of "jogging". This indicates that the given reading is more similar to those recorded during the activity of walking upstairs, leading to its classification as "walk upstairs". This method is effective for classifying data points based on similarity to known examples, and it's particularly useful in contexts like activity recognition from accelerometer data, where patterns of motion can be distinctive but vary slightly from person to person.\end{minipage}\tabularnewline\tabularnewline\end{tabular} \\ \hline
\multirow{3}{*}{\begin{tabular}[c]{@{}l@{}}SimCLR\\ Encoding\end{tabular}} &
  Query &
  \begin{tabular}[c]{@{}l@{}} \begin{minipage}[t]{0.8\columnwidth} The following given embeddings correspond to ``walking upstairs'': {[}{[}14.226081 6.4440064{]}...{[}0.57457817 -22.347515{]}{]} and the following given embeddings correspond to ``jogging'': {[}{[}56.372967   5.0068436{]}...{[}53.95932 -21.747236{]}{]}; classify the embedding {[}14.239516  7.1703763{]} as either ``walking upstairs'' or ``jogging'' considering the minimum distance to the example embeddings provided that the distance metric chosen is euclidean distance. Answer in one word.\end{minipage}\tabularnewline\tabularnewline\end{tabular} \\ \cline{2-3} 
 &
  Response &
  \begin{minipage}[t]{0.8\columnwidth} The embedding {[} 14.239516 , 7.1703763 {]} is classified as ``walking\_upstairs'' based on the minimum distance to the example embeddings, using the Euclidean distance metric. \end{minipage} \\ \cline{2-3} 
 &
  Reasoning &
  \begin{tabular}[c]{@{}l@{}} \begin{minipage}[t]{0.8\columnwidth} To determine the classification of the given embedding{[} 14.239516 , 7.1703763 {]} as either ``walking upstairs'' or ``jogging'', we calculated the Euclidean distance between this embedding and each of the example embeddings provided for both activities. The Euclidean distance metric measures the straight-line distance between two points in Euclidean space. Here's how the decision was made: Calculating Distances: For each category (``walking upstairs'' and ``jogging''), we calculated the distance from the given embedding to all embeddings in that category. Minimum Distance: We then identified the minimum distance from the given embedding to the embeddings in each category. This means we found the closest embedding to the given embedding within both categories. Comparison: The category (``walking upstairs'' or ``jogging'') whose embeddings had the smallest minimum distance to the given embedding is considered the most likely category for that embedding. In this case, the minimum distance between the given embedding and the ``walking upstairs'' embeddings was smaller than the minimum distance to any of the ``jogging'' embeddings. This indicates that the given embedding is more similar to the ``walking upstairs'' activity based on the spatial distribution of embeddings in the feature space. Thus, considering the principle that similar activities will result in embeddings that are closer together in space, the classification was based on the proximity of the given embedding to the known embeddings for each activity, leading to the conclusion that the given embedding is most likely associated with ``walking upstairs''.\end{minipage}\tabularnewline\tabularnewline\end{tabular} \\ \hline
\multirow{3}{*}{\begin{tabular}[c]{@{}l@{}}TFC\\ Encoding\end{tabular}} &
  Query &
  \begin{tabular}[c]{@{}l@{}} \begin{minipage}[t]{0.8\columnwidth} The following given embeddings correspond to ``walking upstairs'': {[}{[}4.256844 -10.712115{]} ... {[}39.49697 -33.193176{]}{]}   and the following given embeddings correspond to ``jogging'': {[}{[}38.99644 -7.460116{]} ... {[}40.007973 4.696819{]}{]}; classify the embedding {[}2.0370352 3.47346{]} as either ``walking upstairs'' or ``jogging'' considering the minimum distance to the example embeddings provided that the distance metric chosen is euclidean distance. Answer in one word.\end{minipage}\tabularnewline\end{tabular} \\ \cline{2-3} 
 &
  Response &
  \begin{tabular}[c]{@{}l@{}} \begin{minipage}[t]{0.8\columnwidth} The embedding {[} 2.0370352, 3.47346 {]} is classified as ``walking\_upstairs'' based on the minimum Euclidean distance to the example embeddings provided for ``walking upstairs'' and ``jogging''. This classification indicates that the given embedding is closer to the set of embeddings associated with ``walking upstairs'' than it is to those associated with ``jogging''.\end{minipage}\tabularnewline\end{tabular} \\ \cline{2-3} 
 &
  Reasoning &
  \begin{tabular}[c]{@{}l@{}} \begin{minipage}[t]{0.8\columnwidth} To determine the classification of the embedding {[} 2.0370352 , 3.47346 {]} {[}2.0370352,3.47346{]}, I calculated the Euclidean distance between this point and each of the given embeddings for ``walking upstairs'' and ``jogging''. After computing the distances to all provided embeddings, I identified the minimum distance from the point to the embeddings of each activity category. The logic behind this approach is that an embedding is more likely to belong to the category to which it is closest in terms of distance in the embedding space. This space is structured so that similar activities (in terms of sensor readings, motions, or whatever features these embeddings represent) are closer together. The results showed that the closest distance between the point {[} 2.0370352 , 3.47346 {]} and the ``walking upstairs'' embeddings was significantly smaller than the closest distance to the ``jogging'' embeddings. This implies that the point is more similar to, or shares more characteristics with, the ``walking upstairs'' activity than with ``jogging''. Therefore, based on the principle of nearest neighbors (where an entity is more likely to be similar to its nearest neighbors in a feature space), the embedding was classified as ``walking upstairs''. This method is a fundamental concept in machine learning for classifying data points based on the known classification of other points nearby in the feature space.\end{minipage}\tabularnewline\end{tabular} \\ \hline
\end{tabular}
\end{table}
\subsection{Time and Cost Analysis of Annotation}
One of the critical factors determining the usage of LLMs as virtual sensors could be the pricing model for the API-based access and the time required to get the annotated data. To assess this, we performed a separate small study with the encoded data from subject 14 of the MotionSense dataset~\cite{malekzadeh2019mobile} for the two classes ``Jogging'' and ``Upstairs'' with a total of 2330 data samples. For the encoding, we used the pre-trained encoder trained using the TFC approach and fixed the dimensions, distance, and number of examples as 2, Euclidean, and 25, respectively. Given this setup and the API access restrictions~\cite{gpt_rate_limit}, it took us $\approx 13.33$ minutes and USD 5.03\$ to get this data annotated using the GPT-4 API.
\subsection{Reasoning behind the Responses}
A critical problem we observed during the motivational experiments (see Section~\ref{motiv_reason}) using a query without contextual information like distance metric and examples was in GPT-4 failing to provide concrete reasoning~\cite{zhao2023explainability} of how it came to a particular conclusion regarding the responses. Notably, redesigning the query and adding some context information, like a proper distance metric, allowed the LLM to provide better reasoning for the responses. A summary of the same is provided in \tablename~\ref{tbl:reason_table}.
\begin{wrapfigure}{L}{0.40\textwidth}
\centering
\includegraphics[width=0.35\textwidth]{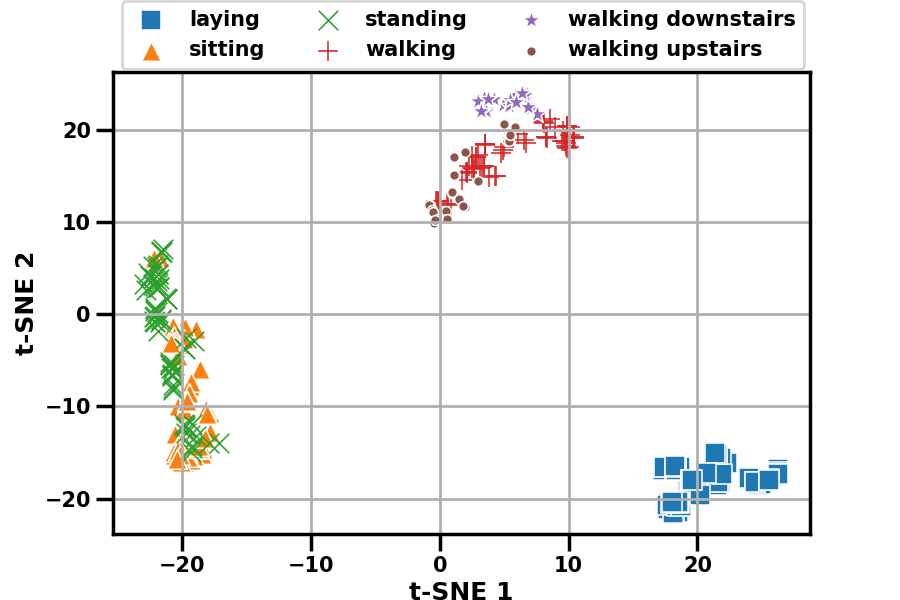}
\caption{t-SNE plots from SimCLR embeddings for the UCI HAR dataset. The embeddings of certain classes, like ``walking'', ``walking upstairs'', and ``walking downstairs'', cluster very closely in the embedding space. This results in a drop in the annotation accuracy provided by GPT-4 for the samples from these closely related classes.}
\label{fig:uci_har_tsne}
\end{wrapfigure}
\subsection{Comprehensive Multi-class Analysis}
After confirming our initial hypothesis with two classes, we next conduct a comprehensive multi-class analysis using the UCI HAR dataset due to its smaller test set size than other datasets. \figurename~\ref{fig:both_figures} provides a detailed report of the overall accuracy we observe with increasing classes. Moreover, as we increased the number of classes, we clearly observed a decline in performance from the LLM. This highlights the importance of identifying classes prone to confusion for deeper insights into this phenomenon, which we will investigate next. From the confusion matrices (shown in \figurename~\ref{fig:conf_mat}), we see that up to three classes, the LLM can still annotate \figurename~\ref{fig:a} with 100\% accuracy. However, when we start adding new classes in the task of annotation, we observe a significant decline in the performance (see \figurename~\ref{fig:b}--\figurename~\ref{fig:d}). One of the key reasons behind this is the inherent confusion introduced by the new classes, like ``walking'', when similar representations are already present for another closely related class, like ``walking upstairs'' and ``walking downstairs''. Similar observation is also observed for classes like ``standing'' and ``sitting''. This can be linked to the overall representational capability of the raw sensor data from the triaxial accelerometer in differentiating between these classes where we see that embeddings for these classes are closely clustered in the embedding space (\figurename~\ref{fig:uci_har_tsne}). Our exploration with both SimCLR and TFC frameworks revealed consistent results across all classes, demonstrating that optimal accuracy was consistently attained with a dimensionality of the test embeddings set to 2.
\subsection{Key Observations}
In summary, the key observations from this study are as follows.
\begin{enumerate}
    \item \textbf{Encoding Raw Signals to Achieve Accurate Annotations:} Encoding the time-series information using self-supervised approaches provides the LLMs with robust representations, which they can then use to generate accurate annotations. Notably, the final obtained embeddings can be reduced significantly to lower dimensions to lower the number of tokens (and the associated cost) while enhancing the accuracy.
    \item \textbf{Adding Context to Query for Accurate and Reasonable Annotations:} One of the key observations that we gain from this study is that even simple information like distance metric adds a significant amount of context to the query. This allows the LLM then to generate an accurate response with better reasoning.
    \item \textbf{Consistency with Higher Number of Examples --  A Cost vs Accuracy Trade-off:} A primary advantage of using SSL-based pre-trained encoders is that they allow similar representations to cluster in the embedding space without the need for label information. This, in turn, facilitated the LLM to provide more accurate responses with increasing examples. However, a critical limitation is that adding more examples to the query can significantly increase the number of tokens and, thus, the cost of getting annotations using the paid APIs.
    \item \textbf{Effects of Increasing the Number of Class for HAR Dataset -- The Choice of the Classes and the Inherent Limitation of the Modality:} The overall representational capability of the data itself limits the approach for annotating the raw sensor data using LLMs. For example, when certain classes are added for the annotation task, the representations obtained using the triaxial accelerometer data become confusing enough for the LLM to clearly differentiate the classes from each other. This highlights that it is not just increasing the class but also the choice of the type of the selected class for annotation and the modality used that impacts the overall accuracy of the annotations obtained from the LLM.
\end{enumerate}

\section{Discussion and Future Work}
\label{discuss}
In this section, we summarise some of the future ideas that can be of interest to the community, given the outcomes of this study.
\begin{enumerate}
    \item \textbf{Virtual Annotation using Vector Databases:} Recent growth in LLMs has also fueled the development of Vector databases~\cite{vector_database}. These databases usually contain high-dimensional embeddings or representations obtained from trained models. Interestingly, the broad idea explored in this paper can be further refined by deploying platforms for annotation-as-a-service where the representations from these databases can assist the system in obtaining a few bootstrap samples for any given physical activity mapping. Subsequently, the user can submit the embeddings obtained from an encoder trained using a self-supervised algorithm for obtaining labels. This would indeed allow the user to get their datasets annotated while not needing to transfer any data to third-party servers or crowdsourcing platforms.
    \item \textbf{Refined Label Space using Active Learning with Annotations from LLM:} Following up on the aforementioned idea of vector databases, a further step towards developing an optimized, high-quality annotation platform can be by applying active learning~\cite{hossain2019active} with sample labels obtained from LLMs. In this case, the first set of bootstrap-labeled data can be provided by the LLM. In contrast, active learning can help refine the label quality for samples that were difficult for the LLM to classify accurately. Interestingly, such an approach may also help reduce hallucinations by opportunistically including labels with human feedback~\cite{hanneke2018actively}.
    \item \textbf{Lack of Annotator Agreement:} One of the critical benefits of the traditional human-in-the-loop annotation approach is in assessing the confusion in labeling through annotator agreement, which, when properly assessed, can significantly impact the decisions made by an ML model~\cite{gordon2022jury}. However, with a more stringent setup for labeling controlled by a single LLM, this approach of assessing the confusion in the obtained label will be significantly hampered. Although a straightforward solution can be using multiple such LLMs and observing their outputs for a given data set, this approach will surely increase the overall annotation cost.
    \item \textbf{Increasing Cost with Multiclass Annotation:} One of the critical restrictions, with the given assessment using the paid API access of GPT-4, would be in the overall cost of annotating the samples. For example, with the number of classes increasing beyond two, the number of examples provided in the prompt will also increase. This, in general, will increase the number of tokens, which in turn will increase the cost significantly.
\end{enumerate}
\section{Conclusion}
\label{conclusion}
Recent LLMs are trained on vast amounts of publicly available data, which makes them a potential replacement for humans in the conventional human-in-the-loop-based annotation scheme. Definitely, this can offer many advantages of being more scalable, timely, and efficient. However, understanding complex time-series data has been a significant challenge for the LLMs, with most of the SOTA approaches proposing either to adapt the computationally expensive ways of fine-tuning the LLM or perform sophisticated prompt-tuning. This paper presents a different perspective on using LLMs for time-series physical sensing data by encoding them with pre-trained time-series encoders. To arrive at this idea and assess it thoroughly, we first segregated the entire design setup into two phases. The first phase was more dedicated to understanding the challenges of using raw sensor data with LLMs. We saw that GPT-4 could not generate accurate and reasonable annotations even with examples by looking at the raw sensor data. This understanding laid the foundation for the next phase, where we systematically looked into providing expert guidance by including distance as contextual information. Then, to exploit the idea of neighborhood and proximity in the embedding space, we studied using SSL-based pre-training of a time-series encoder. These encoders allow the projection of the raw sensor data in the embedding space so that similar classes form distinct clusters. Thus, with examples drawn from a cluster, the next task for the LLM becomes just assessing the proximity. The observations that we gain from the principled evaluation of this approach on four benchmarking datasets validates the broad intuition of using SSL-based time-series encoders as a plug-and-play module over LLMs for virtual annotation.

\bibliographystyle{ACM-Reference-Format}
\bibliography{ref}

\end{document}